\documentclass{article}

\usepackage[pdftex]{graphicx}

\usepackage{cite}
\usepackage[margin=1in]{geometry}
\usepackage{url,dblfloatfix,comment}
\RequirePackage[colorlinks,citecolor=blue,urlcolor=blue]{hyperref}
 \usepackage{float}
\usepackage{amsmath, amsfonts, amssymb, amsthm,color}

\newtheorem{remark}{Remark}

\usepackage{bm}
\usepackage[capitalise]{cleveref}

\usepackage{algorithm, algorithmic}

\usepackage{commath}

\newcommand{\ie}{\emph{i.e.},}

\newcommand{\bV}{\boldsymbol{V}}

\newcommand{\bI}{\boldsymbol{I}}

\newcommand{\bX}{\boldsymbol{X}}
\newcommand{\bA}{\boldsymbol{A}}
\newcommand{\bS}{\boldsymbol{S}}
\newcommand{\bZero}{\boldsymbol{0}}
\newcommand{\bW}{\boldsymbol{W}}

\newcommand{\bw}{\boldsymbol{w}}

\newcommand{\bB}{\boldsymbol{B}}

\newcommand{\br}{\boldsymbol{r}}
\newcommand{\bp}{\boldsymbol{p}}
\newcommand{\bR}{\boldsymbol{R}}
\newcommand{\bc}{\boldsymbol{c}}
\newcommand{\bv}{\boldsymbol{v}}
\newcommand{\bz}{\boldsymbol{z}}

\newcommand{\bx}{\boldsymbol{x}}

\newcommand{\ba}{\boldsymbol{a}}
\newcommand{\bu}{\boldsymbol{u}}

\newcommand{\bQ}{\boldsymbol{Q}}

\newcommand{\by}{\boldsymbol{y}}
\newcommand{\bU}{\boldsymbol{U}}
\newcommand{\bLambda}{\boldsymbol{\Lambda}}

\newcommand{\bepsilon}{\boldsymbol{\epsilon}}

\newcommand{\bSigma}{\boldsymbol{\Sigma}}

\newcommand{\argmax}{\mathop{\rm argmax}}
\newcommand{\argmin}{\mathop{\rm argmin}}

\newcommand{\nn}{\nonumber}
\newcommand{\sspan}{\operatorname{span}}

\providecommand{\norm}[1]{\lVert#1\rVert}

\providecommand{\set}[1]{\left\{#1\right\}}

\providecommand{\bydef}{\overset{\text{def}}{=}}

\usepackage[font=footnotesize]{subfig}

\def \1{\mathbf 1}

\def \R{\mathbb{R}}

\begin{document}
%
\title{Streaming PCA and Subspace Tracking: \\The Missing Data Case}
%
%
%

\author{Laura Balzano\thanks{L. Balzano is with the Department of Electrical Engineering and Computer Science, University of Michigan, Ann Arbor, MI 48109, USA (e-mail: girasole@umich.edu).}~,  
        Yuejie~Chi\thanks{Y. Chi is with the Department of Electrical and Computer Engineering, Carnegie Mellon University, Pittsburgh, PA 15213, USA (e-mail: \mbox{yuejiechi@cmu.edu}).}~, 
        and~Yue~M.~Lu\thanks{Y. M. Lu is with the John A. Paulson School of Engineering and Applied
Sciences, Harvard University, Cambridge, MA 02138, USA (e-mail: yuelu@seas.harvard.edu).}}
\maketitle

\begin{abstract}
For many modern applications in science and engineering, data are collected in a streaming fashion carrying time-varying information, and practitioners need to process them with a limited amount of memory and computational resources in a timely manner for decision making. This often is coupled with the missing data problem, such that only a small fraction of data attributes are observed. These complications impose significant, and unconventional, constraints on the problem of streaming Principal Component Analysis (PCA) and subspace tracking, which is an essential building block for many inference  tasks in signal processing and machine learning. This survey article reviews a variety of classical and recent algorithms for solving this problem with low computational and memory complexities, particularly those applicable in the big data regime with missing data. We illustrate that streaming PCA and subspace tracking algorithms can be understood through algebraic and geometric perspectives, and they need to be adjusted carefully to handle missing data. Both asymptotic and non-asymptotic convergence guarantees are reviewed. Finally, we benchmark the performance of several competitive algorithms in the presence of missing data for both well-conditioned and ill-conditioned systems.

\end{abstract}


%


\section{Introduction}

The explosion of data collection across a variety of domains, for purposes that range from scientific to commercial to policy-oriented, has created a data deluge that requires new tools for extracting useful insights from data. Principal Component Analysis (PCA)  \cite{jolliffe1986pca} and subspace tracking are arguably some of the most commonly used tools for exploring and understanding data. The fundamental mathematics and algorithms for identifying signal subspaces from data have been studied for nearly a century. However, in the modern context, many novel challenges arise, due to the severe mismatch between limited resources available at computational platforms and increasing demand of processing high-dimensional data. In particular, this survey article is motivated by the following aspects of modern data processing.
\begin{itemize}
\item \emph{Large-scale and high-rate.} Data are collected at an extremely large scale with many variables, such as in video surveillance or internet monitoring, and they can accumulate at such high rates that real-time processing is necessary for timely decision making. Therefore, classical batch algorithms for data processing are replaced by online, streaming algorithms that have much smaller memory and computational footprints. 
\item \emph{Missing data.} At each time instance, only a very small subset of the data attributes may be measured, due to hardware limitations, power constraints, privacy concerns, or simple lack of observations. Therefore, classical algorithms that do not account for missing data may yield highly sub-optimal performance and need to be redesigned.
\end{itemize}

To elaborate on these modern challenges, we describe two concrete examples in more detail. First, consider recommendation systems~\cite{sarwar2002incremental}, where users' past product use and opinions are collected. Based on such data, the system attempts to predict other products of interest to those (and potentially other) users. This is of course a scenario involving extremely sparse observations in high dimensions---a user has only purchased or rated a vanishingly small number of products from a company. Moreover, as the users rate more products and as new products become available, it is desirable to update the system's predictions on user preference in an online manner. 

As another example, consider the rigid structure from motion problem in computer vision~\cite{kennedy2016online, tomasi1992shape}. One seeks to build a 3D model of a scene based on a sequence of 2D images that, for an orthographic camera, are projections of that scene onto a plane. Features in the scene can be tracked through the images, and a matrix of their locations in the images has a low-rank (3-dimensional) factorization in terms of the true 3D locations of feature points and the locations of the cameras at each image frame. The problem is obviously high dimensional, and it is also natural to consider the streaming setting, as large numbers of features can be tracked across image frames that arrive sequentially at a high rate. Moreover, many points in the scene are not visible in all image frames due to occlusion. Therefore, while the low-rank subspace of the data recovers the 3D structure of the entire scene, one must estimate this subspace in the presence of missing data.

The list of modern applications continues. The question is: {\em can we have scalable and accurate algorithms for subspace learning that work well even in the presence of missing data in a dynamic environment?}     

\subsection{Subspace Models and Missing Data}
Subspace models have long been an excellent model for capturing intrinsic, low-dimensional structures in large datasets. A celebrated example, PCA \cite{jolliffe1986pca}, has been successfully applied to many signal processing applications including medical imaging, communications, source localization and clutter tracking in radar and sonar, computer
vision for object tracking, system identification, traffic data analysis, and speech recognition, to name just a few. The calculated principal components and best-fit subspaces to a dataset not only allow dimensionality reduction but also provide intermediate means for signal estimation, noise removal, and anomaly detection \cite{scharf1991svd}. As we highlight in this paper, the principal components can be updated using incoming data in a streaming manner, thus offering tracking capabilities that are necessary for real-time decision making.

While there are a plethora of traditional algorithms for performing PCA on a batch dataset 
and for estimating and tracking the principal components in a streaming scenario (see, \emph{e.g.}, \cite{comon1990tracking}, for a survey of earlier literature), most of these algorithms were developed during a time when datasets of interest had a moderate number of variables (say 10-100) and were collected in a controlled environment with little or no missing entries. As argued earlier, modern datasets are being collected on vast scales, in a much less controlled way, often with overwhelmingly many missing entries. In light of this prevalent and modern challenge in signal processing and machine learning, classical algorithms must be adjusted in order to gracefully handle missing data.

\sloppypar When do we have hope to recover missing data? If the complete high-dimensional data are well-approximated by their projection onto a lower-dimensional subspace, and hence in some sense redundant, then it is conceivable that incomplete or subsampled data may provide sufficient information for the recovery of that subspace. A related problem in the batch setting is the celebrated problem of low-rank matrix completion \cite{candes2009exact,chen2018harnessing}, which suggests that it is possible to recover a highly incomplete matrix if its rank is much smaller than the dimension. This is the central intuition that motivates work on \emph{streaming PCA and subspace tracking with missing data}. A burst of research activity has been devoted to developing algorithms and theoretical underpinnings for this problem over the last several years in signal processing, machine learning, and statistics.  
Moreover, powerful results from random matrix theory and stochastic processes have been leveraged to develop performance guarantees for both traditional and newly proposed methods.
At the same time, these methods are also finding new applications to emerging data science applications such as  monitoring of smart infrastructures \cite{gao2016missing}, neurological, and 
physiological signal processing and understanding \cite{majumdar2014low}.

\subsection{Overview of Subspace Tracking Algorithms}\label{sec:overview_methods}
 
There is a long history of subspace tracking algorithms in the literature of signal processing. An extensive survey of methods prior to 1990 was provided in a popular Proceedings of the IEEE article by Comon and Golub \cite{comon1990tracking}. As the common problem dimension was relatively small at that time, the focus was mostly on performance and computational complexity for fully observed data of moderate dimensions. Since then, new algorithms have been and continue to be developed with a focus on minimizing computation and memory complexity for very high-dimensional problems with missing data, without suffering too much on performance \cite{delmas2010subspace}. Consider the problem of estimating or tracking a $k$-dimensional subspace in $\mathbb{R}^{d}$, where $k\ll d$. For modern applications, it is desirable that both the computational complexity (per update) and the memory complexity scale at most linearly with respect to $d$. Moreover, modern applications may require the algorithm to handle a range of missing data, from just a small fraction of missing entries to the information-theoretic limit of only $\mathcal{O}(k \log d)$ entries observed in each data vector\footnote{This is the information-theoretic lower bound of measurements for an arbitrary incoherent rank-$k$ matrix when entries from about $d$ total column vectors are observed uniformly at random \cite{candes2010power}. For a generic matrix, we need only $\mathcal{O}(\max(k, \log d))$ entries per column with $\mathcal{O}(kd)$ total columns \cite{pimentel2016characterization}.}.

Broadly speaking, there are two perspectives from which researchers have developed and studied streaming PCA and subspace tracking algorithms, as categorized by Smith \cite{smith1997subspace}. The first class of algorithms can be interpreted through an algebraic lens; these can be regarded as variants of incremental methods for calculating top-$k$ eigenvectors or singular vectors of a time-varying matrix, such as the sample covariance matrix. Since this time-varying matrix is typically updated by a rank-one modification, various matrix manipulation techniques can be exploited to reduce computational and memory complexities. This viewpoint is particularly useful for understanding algorithms such as Incremental SVD \cite{brand2002incremental}, Karasalo's method \cite{karasalo1986estimating}, Oja's method \cite{oja1982simplified}, Krasulina's method \cite{balsubramani2013fast, krasulina1969method}, and other algorithms based on power iterations  \cite{abed2000fast,hua1999new}, to name a few.

The other class of algorithms can be interpreted through a geometric lens. These algorithms are constructed as the solution to the optimization of a certain loss function, \emph{e.g.}, via gradient descent, designed in either Euclidean space or on a matrix manifold such as the Grassmannian. We focus mainly on methods where the loss function is updated by one additional term per streaming column vector, and the previous estimate can be used as a warm start or initialization. This viewpoint is particularly useful in the presence of missing data, since missing data are easily incorporated into a loss function, and has therefore been leveraged more often than the algebraic viewpoint in the design of subspace tracking algorithms that are tolerant to missing data. Examples include GROUSE \cite{balzano2010online}, PETRELS \cite{chi2013petrels}, ReProCS \cite{guo2014online}, PAST \cite{yang1995projection}, online nuclear norm minimization \cite{mardani2015subspace}, and other algorithms based on stochastic approximation \cite{feng2013online}, to name a few.

The two classes of algorithms, while having distinct features, can often be unified, as an algorithm can often be interpreted through both perspectives. The trade-off between convergence speed in static environments and tracking speed in dynamic environments is also an important consideration in practice,  achieved by balancing the influence from historical data and current data. This can be done by discounting historical data in the construction of the time-varying matrix in algebraic methods, and in the construction of the loss function or selection of step sizes in geometric methods. 

There is also a vast literature on establishing theoretical performance guarantees for various streaming PCA and subspace tracking algorithms. Classical analysis is primarily done in the asymptotic regime (see, \emph{e.g.}, \cite{yang1996asymptotic,chen1998global}), where the discrete-time stochastic processes associated with the algorithms are shown to converge, in the scaling limit \cite{Ljung:1977, Kushner:1977}, to the solution of some deterministic differential equations. Recent developments in performance analysis include new and more tractable asymptotic analysis for high-dimensional cases \cite{WangL:16, WangEL:17, WangML:17, WangEL:18}, as well as finite-sample probabilistic performance guarantees \cite{mitliagkas2013memory, hardt2014noisy, li2018near, shamir2016convergence, allen2016first}.

\subsection{Organization of the Paper}

We first describe in~\cref{sec:formulation} the problem formulation of PCA and streaming PCA in the presence of missing data.
We then survey algorithms that perform streaming subspace estimation and tracking with full or incompletely observed columns:  
\cref{sec:algebraic_methods} focuses on those using algebraic approaches and \cref{sec:geometric} on those using geometric approaches.   
Many of these algorithms have associated theoretical analysis with regards to the estimation accuracy and algorithmic convergence rates, which we discuss in \cref{sec:analysis}. 
We then provide numerical comparisons of a number of competitive algorithms in \cref{sec:comparisons} and conclude in~\cref{sec:final}.  

\subsection{Notations}

Throughout this paper, we use boldface letters to denote vectors and matrices, \emph{e.g.}, $\ba$ and $\bA$. For a positive semidefinite (PSD) matrix $\bA$, we write $\bA \succeq 0$. The transpose of $\bA$ is denoted by $\bA^{T}$, and $\| \bA\|$, $\|\bA\|_{\mathrm{F}}$, and $\mathrm{Tr}(\bA)$ denote the spectral norm, the Frobenius norm and the trace, respectively.  The expectation of a random variable $a$ is written as $\mathbb{E}[a]$. The identity matrix of dimension $k$ is written as $\bI_k$. We shall use $d$ to denote the dimension of the fully observed data vector and $k$ to denote the dimension of the subspace to be estimated. A subscript $n$ on the data vector $\bx_n \in \R^d$ refers to its order in a sequence of vectors, and the notation $\bx_n(i)$ refers to the $i$th component of the vector $\bx_n$. 



\section{Problem Formulation}\label{sec:formulation}
 
In this section, we will start by formulating the problem of subspace estimation in the batch setting, which serves as a good starting point to motivate streaming PCA and subspace tracking in the streaming setting with missing data.

\subsection{PCA in the Batch Setting} \label{sec:batch}

The PCA or subspace estimation problem can be formulated either \emph{probabilistically}, where data are assumed to be random vectors drawn from a distribution with mean zero and some covariance matrix whose principal subspace we wish to estimate, or \emph{deterministically}, where we seek the best rank-$k$ subspace that fits the given data. Both models are used extensively throughout the literature. The former is used more prevalently in the signal processing and statistics literature, while the latter is more prevalent in applied mathematics, optimization, and computer science literature. The problem formulations result in equivalent optimization problems, and so we put them here together for a unified view.

(a) \emph{Probabilistic view}: Consider a stationary, $d$-dimensional random process $\bx \in \mathbb{R}^d$, which has a zero mean and a covariance matrix $\bSigma = \mathbb{E}[\bx\bx^T]$.\footnote{It is straightforward to consider the complex-valued case $\bm{x} \in \mathbb{C}^d$, but we only consider the real case in this survey for simplicity. For more information, see \cite{golub2012matrix}.} Denote the eigenvalue decomposition (EVD) of the covariance matrix as $\bSigma=\widetilde{\bU}\bLambda\widetilde{\bU}^T$, where $\widetilde{\bU}=[\bu_1,\ldots,\bu_d]$ has orthonormal columns, and $\bLambda=\mbox{diag}\{\lambda_1,\ldots,\lambda_d\}$, where $\lambda_1\geq\lambda_2\geq\cdots\geq\lambda_d\geq 0$ are the eigenvalues arranged in a non-increasing order. Our goal is to estimate the top-$k$ eigenvectors, also called the principal components $\bU^{\ast}=[\bu_1,\ldots,\bu_k]$, of $\bSigma$, given a finite number of i.i.d. data samples, $\bx_1,\bx_2,\ldots, \bx_n \sim \bx$. Note that we do not require $\bSigma$ be a rank-$k$ matrix.

(b) \emph{Deterministic view}: In a deterministic formulation, the data samples $\bx_1, \dots, \bx_n \in\mathbb{R}^d$ are considered arbitrary. We wish to find the rank-$k$ subspace that best fits these data in the sense of minimizing the $\ell_2$ projection error, that is 
\begin{align} 
\widehat{\bU}_n &= \argmin_{\bU\in\mathbb{R}^{d\times k}, \bU^T\bU=\bI_k}  \sum_{\ell=1}^n \left\| \bx_\ell - \mathcal{P}_{\bU}(\bx_\ell) \right\|_2^2 \label{deterministic_PCA} \\
& =  \argmin_{\bU\in\mathbb{R}^{d\times k}, \bU^T\bU=\bI_k}  \left\|\bX_n - \bU\bU^T\bX_n \right\|_{\mathrm{F}}^2 \; \label{def_SVD} \\
& = \argmax_{\bU\in\mathbb{R}^{d\times k}, \bU^T\bU=\bI_k} \mbox{Tr}\left(\bU\bU^T\bSigma_n\right),\label{def_PCA}
\end{align}
where $\mathcal{P}_{\bU}$ denotes the projection operator onto the column span of the matrix $\bU$ and $\mathcal{P}_{\bU}=\bU\bU^T$ when $\bU$ has orthonormal columns, $\bX_n=[\bx_1,\bx_2,\ldots, \bx_n]$ concatenates the data vectors as columns into a matrix, and  $\bSigma_n =  \sum_{\ell=1}^n\bx_\ell \bx_\ell^T =  \bX_n\bX_n^T$ is the (unscaled) Sample Covariance Matrix (SCM). 

The equivalence of \eqref{def_SVD} and \eqref{def_PCA} suggests that finding the subspace that maximizes the explained variance of $\bSigma_n$ is equivalent to minimizing the approximation error of the data matrix $\bX_n$. While the formulations \eqref{def_SVD} or \eqref{def_PCA} are non-convex, both due to the cost function's non-convexity in $\bU$ and the non-convex constraint $\bU^T\bU = \bI_k$, they admit a well-defined solution, solved by the SVD of $\bX_n$, equivalently the EVD of $\bSigma_n$, as was discovered independently by \cite{schmidt1907theorie} (see \cite{stewart1993early,stewart2011FHS} for details) and \cite{eckart1936approximation}. Specifically, the solution $\widehat{\bU}_n$ is given as the top-$k$ eigenvectors of the SCM $\bSigma_n$.

(c) \emph{Unified perspective}:
Consider the following expected loss function
\begin{equation}\label{population_loss}
J(\bU) = \mathbb{E}\,\| \bx - \bU\bU^T\bx\|_2^2, 
\end{equation}
where $\bU\in\mathbb{R}^{d\times k}$ and the expectation is taken with respect to $\bx$ which is zero mean with a covariance matrix $\bSigma$. The following important result was proven in \cite{yang1995projection}:  if $\lambda_k>\lambda_{k+1}$, \emph{i.e.}, if there is a strict eigengap, then the global optima of $J(\bU)$ correspond to $\bU$ that contains the top-$k$ eigenvectors of $\bSigma$ up to an orthonormal transformation, matching the solution of PCA in the probabilistic view. Interestingly, the solution to the deterministic formulation \eqref{deterministic_PCA} can be thought of as an empirical version of \eqref{population_loss}, if the data samples are indeed drawn according to the probabilistic model. Moreover, in this case, $\widehat{\bU}_n$ produces an order-wise near-optimal estimate to $\bU^{\ast}$ for a large family of distributions \cite{vershynin2012close}. In this regard, the two formulations are equivalent in some sense, though in the deterministic setting, there need not be any generative model or ``ground truth'' for the underlying subspace.

\subsection{Streaming PCA and Subspace Tracking}
\label{sec:missing}

In a streaming setting, the data samples arrive sequentially over time, with each sample only seen once,\footnote{This is different from what is known as a stochastic setting, where samples may be accessed at multiple times or in multiple passes.} and one wishes to update the subspace estimate sequentially without accessing historical data. In a dynamic environment, either the covariance matrix or the best rank-$k$ subspace can be time-varying --- therefore, we wish to  track such changes as quickly as possible. 

 In this survey article, we use the terminology ``streaming PCA" and ``subspace tracking" interchangeably to refer to algorithms that can update and track a data subspace using streaming observations. Nonetheless, we acknowledge they have different connotations and indeed they have arisen from different contexts. The terminology ``subspace tracking'' is common in the literature of signal processing \cite{comon1990tracking}, where one often needs to update the subspace in a dynamic environment as in array signal processing or communications. The more recent terminology of ``online PCA" or ``streaming PCA'' can be found in the machine learning literature, motivated by the study in computer science of trying to replicate the behavior of batch PCA with streaming data \cite{muthukrishnan2005data} or data too large for memory. In addition, ``incremental SVD" \cite{brand2002incremental} or ``updating the SVD" \cite{bunch1978updating} are terminology used more classically in numerical methods. It turns out that all of the algorithms reviewed herein can handle both the settings where the underlying subspace is static or time-varying by adjusting parameters within the algorithm such as data discounting factors or step sizes.

Streaming PCA can be considered as a nonconvex stochastic approximation problem, given by \eqref{population_loss}.
The solution to the batch problem that we outlined in Section~\ref{sec:batch} is no longer appropriate for the streaming setting --- it requires one to formulate and store the SCM $\bSigma_n$, which has a memory complexity of $\mathcal{O}(d^2)$, and to estimate the top-$k$ eigenvectors directly from the SCM, which has a time complexity of $\mathcal{O}(nd^2)$. Both these memory and time complexities are too expensive for large-scale problems. It is greatly desirable to have algorithms with computation and memory complexities that grow at most linearly in $d$.

\subsection{Missing Data}
An important setting that we will consider in this survey is {\em missing data}, where only a subset of the coordinates of of each sample $\bx_n$ are observed. 
We denote this measurement as
\begin{equation}
\label{eq:mdoperator}
\mathcal{P}_{\Omega_n}(\bx_n),
\end{equation}
where $\mathcal{P}_{\Omega_n}$ is a projection operator onto the coordinates represented by an observation mask, $\Omega_n\in\{0,1\}^d$, where $\bx_n(i)$ (the $i$th entry of $\bx_n$) is observed if and only if $\Omega_n(i)=1$. This issue poses severe challenges for most PCA algorithms, particularly when the number of observed entries is much smaller than $d$. To begin, one may be concerned with {\em identifiability}: can we find a unique subspace of rank-$k$ that is consistent with the partial observations? Luckily, the answer to this question is yes, at least in the batch setting where the problem is equivalent to that of low-rank matrix completion: under mild assumptions, the low-rank subspace can be reconstructed from subsampled column vectors as long as there are enough observations.\footnote{``Enough" observations per vector here will depend both on the number of vectors and the conditioning or coherence \cite{candes2009exact} of the subspace.} It may also be tempting to execute subspace tracking algorithms by ignoring the missing data and padding with zeros at the missing entries, however the sample covariance matrix constructed in this way leads to a biased estimator \cite{chi2013nearest,yang2015sparse}. Therefore, one must think more carefully about how to handle missing data in this context.


\section{Algebraic Methods}
\label{sec:algebraic_methods}

In this section and the next, we will discuss two classes of algorithms based on algebraic approaches and geometric approaches respectively, as outlined in Section~\ref{sec:overview_methods}. The algebraic approaches are based on finding the top eigenvectors of a recursively updated SCM, or a surrogate of it, given as
\begin{equation}\label{eq:recursive_scm}
\bSigma_n =  \alpha_n \bSigma_{n-1} + \beta_n \bx_n\bx_n^T,
\end{equation}
where $\alpha_n$ and $\beta_n$ balance the contributions from the previous SCM and the current data sample. Two popular choices are equal weights on all time slots, which is
$$ \alpha_n =1 , \quad \beta_n = 1;$$
and discounting on historical data, which is
$$ \alpha_n = \lambda , \quad \beta_n = 1, \quad 0\ll \lambda <1 .$$ 
Equivalently, the above can be reworded as finding the top singular vectors of a recursively updated data matrix $\bX_n$. As we are interested in calculating or approximating the top-$k$ eigenvectors of $\bSigma_n$, algebraic methods use matrix manipulations and exploit the simplicity of the rank-one update to reduce complexity.


\subsection{Incremental Singular Value Decomposition (ISVD)}
\label{sec:isvd}

We begin by discussing the ISVD approach of Bunch and Neilsen \cite{bunch1978updating}, which is an exact method to compute the full SVD of a streaming full data matrix, \ie~with sequentially arriving, full data vectors. This algorithm is given in \cref{alg:isvd} and is the result of some simple observations about the relationship of the SVD of $\bX_n$ and that of $\bX_{n-1}$. Suppose we are given the compact SVD of the data matrix at time $n-1$, 
$$ \bX_{n-1} = \left[ \begin{matrix} \bx_1 & \dots & \bx_{n-1} \end{matrix}\right] = \widetilde \bU_{n-1}  \widetilde \bS_{n-1}  \widetilde \bV_{n-1}^T,$$ 
where $\widetilde \bU_{n-1} \in \R^{d\times d}$ and $\widetilde \bV_{n-1} \in \R^{(n-1) \times (n-1)}$ are orthonormal, and $\widetilde \bS_{n-1} \in \R^{d \times (n-1)}$ is the concatenation of two matrices: a diagonal matrix (of size $\min \{d,n-1\}$) with non-negative non-increasing diagonal entries, and an all-zero matrix. Note that we are using the $\widetilde \bU$ notation for the square orthogonal matrix as opposed to $\bU$ for a $d\times k$ matrix, as in \cref{sec:batch},  
because we are computing the full SVD, not a low-rank approximation. For simplicity of exposition, let's assume $d \geq n$ (but both cases $d<n$ and $d\geq n$ are described in \cref{alg:isvd}). We wish to compute the SVD of 
$$\bX_n = \left[ \begin{matrix}  \bX_{n-1} & \bx_n\end{matrix}\right] = \widetilde{\bU}_{n} \widetilde{\bS}_{n} \widetilde{\bV}_{n}^T,$$
where $\widetilde{\bU}_{n}$, $\widetilde{\bS}_{n}$, and $\widetilde{\bV}_{n}$ are defined similarly as $\widetilde{\bU}_{n-1}$, $\widetilde{\bS}_{n-1}$, and $\widetilde{\bV}_{n-1}$. 

Recognizing that 
$$\bX_{n}\bX_n^T =   \bX_{n-1}   \bX_{n-1}^T + \bx_n \bx_n^T$$ 
and  
$$\widetilde{\bU}_{n-1}^T \bX_n\bX_n^T \widetilde{\bU}_{n-1} = \widetilde{\bS}_{n-1} \widetilde{\bS}_{n-1}^T + \bz_n\bz_n^T $$
where $\bz_n = \widetilde{\bU}_{n-1}^T \bx_n$, we can compute the new singular values by finding the eigenvalues of $\widetilde{\bS}_{n-1} \widetilde{\bS}_{n-1}^T + \bz_n\bz_n^T$ using the zeros of the characteristic equation \cite{golub1973some}, which in this case has a special structure; 
in particular, if $\widetilde \sigma_i$ are the diagonal values of $\widetilde{\bS}_{n-1}$, then the zeros of 
\begin{equation}
\label{eq:isvdsval}
1 + \sum_{i=1}^{d} \frac{z_n(i)^2}{\widetilde\sigma_i^2 - \lambda}
\end{equation} with respect to the variable $\lambda$ identify the eigenvalues of $\widetilde{\bS}_{n-1} \widetilde{\bS}_{n-1}^T + \bz_n\bz_n^T$. Denote the resulting eigenvalues as $\lambda_i$ for $i=1,\dots,d$. To update the left singular vectors to the new $\widetilde{\bU}_n = \left[ \begin{matrix} \bu_1 & \dots & \bu_d \end{matrix} \right]$, we need to solve $\left(\widetilde{\bS}_{n-1} \widetilde{\bS}_{n-1}^T + \bz_n\bz_n^T\right) \bu_i = \lambda_i \bu_i$ and normalize the solution. Therefore \cite{golub1973some, comon1990tracking}, 
\begin{equation}
\label{eq:isvdsvec}
\bu_i = \frac{ (\widetilde{\bS}_{n-1} \widetilde{\bS}_{n-1}^T -\lambda_i \bI  )^{-1}\bz_{n}}{\left\| (\widetilde{\bS}_{n-1} \widetilde{\bS}_{n-1}^T -\lambda_i \bI  )^{-1}\bz_{n}\right\|}, \quad i=1,\ldots, d.
\end{equation} %

 \begin{algorithm}[ht]
\caption{ISVD} \label{alg:isvd}
\begin{algorithmic}[1] 
\STATE{Given $\bx_1$, set $\widetilde \bU_1 = \bx_1/\|\bx_1\|$, $\widetilde \bS_1 = \|\bx_1\|$, $\widetilde \bV_1 = 1$;}
\STATE{ Set $n=2$;}
\REPEAT
\STATE{Define $\bw_n := \widetilde \bU_{n-1}^T \bx_n$;} 
\STATE{Define $\bp_n := \widetilde \bU_{n-1} \bw_n$; $\br_n := \bx_n - \bp_n$;}
\IF{$\|\br_n\| \neq 0$}
\STATE{Compute the SVD of the update matrix:}
\begin{equation}
\left[ \begin{matrix} \widetilde \bS_{n-1} & \bw_n \\ \bZero & \|\br_n\| \end{matrix}
  \right]  =
\widehat{\bU} \widehat{\bS} \widehat{\bV}^T,
\label{eq:update1}
\end{equation} 
by solving \eqref{eq:isvdsval} for $\lambda$, which gives the diagonal entries of $\widehat{\bS}$, where $\widetilde \sigma_i^2$ are the diagonal entries of $\widetilde \bS_{n-1}$  and 
$\bz_n = \left[ \begin{matrix} \bw_n^T & \|\br_n\| & \bZero \end{matrix} \right]^T$,
and then solving \eqref{eq:isvdsvec} for $\widehat \bU$. 
\STATE Set
\begin{align}
\widetilde \bU_{n} &:=  \left[ \begin{matrix} \widetilde \bU_{n-1} & \frac{\br_n}{\|\br_n\|} \end{matrix} \right] \widehat{\bU}, \quad
\widetilde \bS_{n} := \widehat{\bS}\;. \label{eq:isvd-u-update}  \\
\widetilde \bV_{n} &:=  \left[ \begin{matrix} \widetilde\bV_{n-1} &\bZero \\ 0 & 1 \end{matrix} \right] \widehat{\bV}.  \nn
\end{align}
\ELSIF{$\|\br_n\| = 0$}
\STATE{(this happens when $n>d$ or $\bx_n \in \sspan\{\bU_{n-1}\}$)}
\STATE{Compute the SVD of the update matrix:}
\begin{equation}
\left[ \begin{matrix} \widetilde \bS_{n-1} & \bw_n \end{matrix}
  \right]  =
\widehat{\bU} \widehat{\bS} \widehat{\bV}^T,
\label{eq:update2}
\end{equation} 
by solving \eqref{eq:isvdsval} for $\lambda$, which gives the diagonal entries of $\widehat{\bS}$, where $\widetilde \sigma_i^2$ are the diagonal entries of $\widetilde \bS_{n-1}$  and  
$\bz_n = \left[ \begin{matrix} \bw_n^T & \bZero \end{matrix} \right]^T$, 
and then solving \eqref{eq:isvdsvec} for $\widehat \bU$. 
\STATE Set
\begin{align*}
\widetilde \bU_{n} &:=  \widetilde \bU_{n-1}  \widehat{\bU}, \quad
\widetilde \bS_{n} := \widehat{\bS}. \\
\widetilde \bV_{n} &:=  \left[ \begin{matrix} \widetilde\bV_{n-1} &\bZero \\ 0 & 1 \end{matrix} \right] \widehat{\bV}.
\end{align*}
\ENDIF
\STATE{$n:=n+1$;}
\UNTIL{termination}
\end{algorithmic}
\end{algorithm}


So far, the above derivations assume $\widetilde{\bU}_{n-1}$ is a square orthonormal matrix, and the resulting computations are suitable for incremental updates of the full SVD. However, this still requires $\mathcal{O}(dn^2+n^3)$ complexity\footnote{This complexity is assuming $d>n$. In this case, the bottleneck is updating $\widetilde \bU_{n}$ in \eqref{eq:isvd-u-update}, which needs $\mathcal{O}(dn^2)$ operations for a na\"ive matrix multiplication; then to update $\widetilde \bV_n$, we require $\mathcal{O}(n^3)$ operations for the same step.} for {\em every iteration} computing the full SVD (\ie~all singular values and left/right singular vectors) using \cref{alg:isvd}, and the memory requirement grows as $n$ grows if the data are full rank (or low rank with even a very small amount of additive noise), which are both undesirable. 

On the other hand, estimating a thin SVD or the top $k$-dimensional singular subspace can improve computation. In fact, if $\bX_{n-1}$ is exactly rank-$k$, this incremental approach requires fewer computations as pointed out in \cite{brand2006fast}. 
In this case, take the first $k$ columns of $\widetilde \bU_{n-1}$ and call these $\bU_{n-1}$. We only need these to represent $\bX_{n-1}$ because the others correspond to zero singular values. Let $\bS_{n-1}$ and $\bV_{n-1}$ be the corresponding matrices for this thin SVD so that $\bX_{n-1} =  \bU_{n-1} \bS_{n-1}  \bV_{n-1}^T$. We then notice as in \cite{brand2006fast, balzano2013grouse} that 
\begin{align}\label{eq:inc_decomp}
\bX_n &= \left[ \begin{matrix} \bU_{n-1} & \frac{\br_n}{\|\br_n\|} \end{matrix} \right] \left[ \begin{matrix} \bS_{n-1} & \bw_n \\ 0 & \|\br_n \| \end{matrix} \right] \left[ \begin{matrix}  \bV_{n-1}^T & 0 \\ 0 & 1 \end{matrix} \right] 
\end{align} 
where $\bw_n =  \bU_{n-1}^T \bx_n$ are the projection weights onto the span of this now tall matrix $ \bU_{n-1}$ and $\br_n = \bx_n -  \bU_{n-1} \bU_{n-1}^T \bx_n$ is the residual from the projection. 
%
We only must diagonalize the center matrix of \eqref{eq:inc_decomp} to find the SVD of  $\bX_n$.
We only assumed $ \bX_{n-1}$ is rank-$k$, but then to make this assumption at every step is a strong assumption and means the matrix is exactly low-rank. However, this technique is used successfully as a heuristic, by truncating smaller singular values and corresponding singular vectors, even when this assumption does not hold. This method is described in the following section. Finally, we point out that Karasalo's subspace averaging algorithm \cite{karasalo1986estimating} is similar to ISVD, but it uses specific information about the noise covariance.

\subsection{MD-ISVD, Brand's Algorithm and PIMC}
\label{sec:mdisvd}
A major drawback of many linear algebraic techniques is their inapplicability to datasets with missing data. While it is not straightforward how to adapt the ISVD for missing data, there are several different approaches in the literature \cite{brand2002incremental, kennedy2014online, kennedy2016online}. All approaches begin in the same way, by considering how to compute two key quantities, the projection weights $\bw_n$ and the residual $\br_n$, given missing data. Whereas for complete data, we have $\bw_n = \argmin_{\bw} \| \bx_n - \bU_{n-1} \bw\|_2^2 = \bU_{n-1}^T \bx_n$, for missing data one may solve 
\begin{equation}
\label{eq:mdweights}
\bw_n = \argmin_{\bw} \| \mathcal{P}_{\Omega_n} \left(\bx_n - \bU_{n-1} \bw \right) \|_2^2
\end{equation}
where $\mathcal{P}_{\Omega_n}$ is the measurement operator with missing data as in \eqref{eq:mdoperator}. 
Then letting $\bp_n = \bU_{n-1} \bw_n$, define the residual to be 
\begin{equation}
\label{eq:mdresid}
\br_n(i) = \left\{ \begin{matrix} \bx_{n}(i) - \bp_{n}(i) & \text{if}\; \Omega_n(i) = 1 \\ 0 & \text{otherwise} \end{matrix}\right. \;.
\end{equation}
All the methods we describe in this section use these quantities in place of $\bw_n$ and $\br_n$ in \cref{alg:isvd}. They also all mimic \eqref{eq:update1} for the update, and once they have observed enough vectors they \emph{truncate} the $k+1$ singular value and corresponding singular vectors. 

The methods diverge only in the way they replace the singular value matrix $\bS_{n-1}$ in \eqref{eq:update1}. Brand \cite{brand2002incremental} replaces $\bS_{n-1}$ with $\lambda \bS_{n-1}$ where $0\ll \lambda < 1$ is a scalar 
weight that diminishes the influence of previous singular values. If one takes $\lambda = 1$, this is arguably the most direct extension of ISVD to missing data, and following \cite{kennedy2016online} we call this algorithm Missing Data-ISVD (MD-ISVD). Kennedy et~al. \cite{kennedy2014online} present a Polar Incremental Matrix Completion (PIMC) method, which weights $\bS_{n-1}$ with a scalar based on the norm of the data observed thus far. These are different approaches to modeling the uncertainty in the singular values arising from incomplete observations. These algorithms are together summarized in \cref{alg:mdisvd}. These different rules provide different trade-offs in \eqref{eq:recursive_scm}, as they represent different weighting schemes on historical data.

\begin{algorithm}[h]
\caption{MD-ISVD, Brand's algorithm, PIMC} \label{alg:mdisvd}
\begin{algorithmic}[1]
\STATE{Given an orthonormal matrix $\bU_0\in \mathbb{R}^{d \times k}$, $\bS_0 = 0$;}
\STATE{For PIMC, $\gamma_0=1$.}
\STATE{ Set $n=1$;}
\REPEAT
\STATE{Define $\bw_n := \arg \min_{\bw} \|\mathcal{P}_{\Omega_n} \left(\bU_{n-1} \bw - \bx_n \right) \|_2^2$;} 
\STATE{Define $\bp_n := \bU_{n-1} \bw_n$; \begin{equation*}
\br_n(i) = \left\{ \begin{matrix} \bx_n(i) - \bp_n(i) & \Omega_n(i)=1 \\ 0 & \text{otherwise} \end{matrix}\right. \;.
\end{equation*}}
\STATE{Compute the SVD of the update matrix:}
\begin{equation}
\left[ \begin{matrix} \mathbf{\Gamma}_{n-1} & \bw_n \\ \bZero & \|\br_n\| \end{matrix}
  \right]  =
\widehat{\bU} \widehat{\bS} \widehat{\bV}^T,
\label{eq:update_isvd_md}
\end{equation} 
where
\begin{align*}
\mathbf{\Gamma}_{n-1} & = \left\{ \begin{array}{cc}
\bS_{n-1} & \quad \mbox{for MD-ISVD}, \\
\lambda\bS_{n-1}   & \quad \mbox{for Brand's algorithm}, \\
\frac{\gamma_n}{\|\bS_{n-1}\|_{\mathrm{F}}} \bS_{n-1}, & \mbox{for PIMC}, 
\end{array} \right.
\end{align*}
with $\gamma_n^2 = \gamma_{n-1}^2 + \|\mathcal{P}_{\Omega_n} \left(\bx_n \right) \|_2^2 $ for PIMC.
\STATE Set
$\check \bU_{n} :=  \left[ \begin{matrix} \bU_{n-1} & \frac{\br_n}{\|\br_n\|} \end{matrix} \right] \widehat{\bU}$;  
\STATE Set $\bU_n$ as the first $k$ columns of $\check \bU_{n}$ and $\bS_n$ as the top $k$-by-$k$ block of $\widehat{\bS}$.
\STATE{$ n :=n+1$;}
\UNTIL{termination}
\end{algorithmic}
\end{algorithm}


\subsection{Oja's method}

Oja's method was originally proposed in 1982 \cite{oja1982simplified}. It is a very popular method for streaming PCA, and recent attention has yielded significant insight into its practical performance (see discussion in Section \ref{sec:analysis}). Given an orthonormal initialization $\bU_0\in\mathbb{R}^{d\times k}$, at the $n$th time Oja's method updates to $\bU_n$ according to the input data $\bx_n$ as
\begin{equation}\label{Oja}
\bU_n =\Pi( \bU_{n-1} + \eta_n \bx_n\bx_n^T\bU_{n-1} ),  
\end{equation}
where $\Pi(\bW)=\bQ$ is an orthogonalization operator, 
\emph{i.e.},~$\bW=\bQ\bR$ is the QR decomposition. The parameter $\eta_n$ is the step size or learning rate that may change with time.

While Oja's method has not been derived for the missing data case in the literature, following our discussion on ISVD, one realizes that if as before we let $\bw_n = \bU_{n-1}^T\bx_n$ be the coefficient of $\bx_n$ in the previous estimate $\bU_{n-1}$, then Oja's method is equivalent to 
\begin{equation}\label{Oja_rewrite}
\bU_n =\Pi( \bU_{n-1} + \eta_n \bx_n\bw_n^T ).
\end{equation}
A straightforward extension in the missing data case is then to estimate the coefficient $\bw_n$ as \eqref{eq:mdweights}, and to fill in the missing entries in $\bx_n$ as follows. Let $\bp_n = \bU_{n-1} \bw_n$, and the data vector can be interpolated as 
$$\tilde \bx_n  = \left\{ \begin{matrix} \bx_n(i) & \mbox{if}\; \Omega_n(i) = 1 \\ \bp_n(i) & \text{otherwise} \end{matrix}\right.\;.$$
Then Oja's update rule in the missing data case becomes
\begin{equation} \label{eq:ojamd}
\bU_n =\Pi( \bU_{n-1} + \eta_n \tilde \bx_n\bw_n^T ) \;.
\end{equation}
This algorithm is summarized in Algorithm~\ref{alg:Oja}. Note that Oja's original method with full data becomes a special case of this update. We study this extension in the numerical experiments reported in Section~\ref{sec:comparisons}. 
\begin{algorithm}[h]
\caption{Oja's algorithm with missing data}
\begin{algorithmic}[1]
\STATE{Given an orthonormal matrix $\bU_0\in \mathbb{R}^{d \times k}$;}
\STATE{Set $n:=1$;}
\REPEAT 
\STATE{Define $ {\bw}_n :=  \argmin_{\bw} \|\mathcal P_{\Omega_n} \left(   \bx_n  - \bU_{n-1} \bw\right)\|_2^2$ ;} 
\STATE Set $\bp_n = \bU_{n-1} \bw_n$. 
\STATE Set $\tilde \bx_n  = \left\{ \begin{matrix} \bx_n(i) & \mbox{if}\; \Omega_n(i) = 1  \\ \bp_n(i) & \text{otherwise} \end{matrix}\right.\;$
\STATE $\bU_n =\Pi( \bU_{n-1} + \eta_n \tilde{\bx}_n\bw_n^T )$,  
\STATE{$n:=n+1$;}
\UNTIL{termination}
\end{algorithmic}
\label{alg:Oja}
\end{algorithm}

Finally, we note that closely related to Oja's is another method called Krasulina's algorithm \cite{krasulina1969method}, which is developed for updating a rank-$1$ subspace with full data:
\begin{equation}\label{Krasulina}
\bU_n = \bU_{n-1} + \eta_n \left( \bx_n\bx_n^T  - \frac{\bU_{n-1}^T\bx_n\bx_n^T\bU_{n-1}}{\|\bU_{n-1}\|^2}\bI_d\right)\bU_{n-1}.
\end{equation}
It can be viewed as a stochastic gradient descent method with the Rayleigh quotient as its objective. Oja's method is equivalent to Krasulina's method up to the second order terms \cite{oja1985stochastic,balsubramani2013fast}.

\begin{remark}[Block Power Method]
A block variant of Oja's method has been developed in the literature \cite{mitliagkas2013memory,hardt2014noisy,balcan2016improved}, where it partitions the input into blocks and each time processes one block in a way similar to Oja's method. These methods are referred to as the block power method, or block Oja's method, or the noisy power method. They are easier to analyze but yield suboptimal performance \cite{allen2016first}. 
\end{remark}


\section{Geometric Methods}
\label{sec:geometric}

In this section, we review subspace tracking algorithms developed via geometric approaches. These are developed by optimizing certain loss functions over $d \times k$ matrices in Euclidean space or the Grassmann manifold of rank-$k$ subspaces in $\R^d$. Subspace tracking is enabled by optimizing a recursively updated loss function, such as the squared projection loss onto the subspace, as
\begin{equation}
\label{eq:geomloss}
F_n(\bU) =\alpha_n F_{n-1}(\bU) + \beta_n \left\| \bx_n - \mathcal{P}_{\bU}(\bx_n) \right\|_2^2, 
\end{equation}
where $\bU\in\mathbb{R}^{d\times k}$, and $n$ is the time index, which is typically updated by using the previous estimate as a warm start. Similarly, the choice of $\alpha_n$ and $\beta_n$ balances the convergence rate (how fast it converges with data from a static subspace) and the tracking capability (how fast it can adapt to changes in the subspace). 
Additionally, the step size of some gradient algorithms can also be used as a tuning knob for tracking; a more aggressive step size will adapt more quickly to new data. 
Given the necessity of scalable and memory-efficient algorithms, first-order and second-order stochastic gradient descent \cite{bottou2010large} are gaining a lot of popularity recently in signal processing and machine learning.  


\subsection{GROUSE}
\label{sec:grouse}
Grassmannian Rank-One Update Subspace Estimation (GROUSE) was first introduced in~\cite{balzano2010online} as an incremental gradient algorithm to build high quality subspace estimates from very sparsely sampled vectors, and has since been analyzed with fully sampled data \cite{balzano2015local, zhang2016global}, noisy data \cite{zhang2016global}, and missing or compressed data \cite{balzano2015local, zhang2016convergence} (see Section \ref{sec:analysis}). 
The objective function for the algorithm is given by
\begin{equation}
F_n(\bU) = \sum_{\ell=1}^n \|\mathcal P_{\Omega_\ell} (\bx_\ell - \bU\bU^T \bx_\ell)\|_2^2,
\end{equation}
which is a special case of \eqref{eq:geomloss} with $\alpha_n = \beta_n = 1$. 
GROUSE implements a first-order incremental gradient procedure \cite{bertsekas2011incremental} to minimize this objective with respect to the subspace variable $\bU$ constrained to the Grassmannian \cite{edelman1998geometry}, the manifold of all subspaces with a fixed rank, given as
$$ \min_{\bU\in\mathbb{R}^{n\times k}:\bU^T\bU=\bI_k} F_n(\bU).$$ 
GROUSE has iteration complexity $\mathcal{O}(dk + |\Omega_n|k^2)$ at the $n$th update and so is scalable to very high-dimensional applications. 
The algorithm steps are described in \cref{alg:grouse}. 

The GROUSE update in \eqref{eq:gpupdate} can also be written as:

$$\bU_n = \bU_{n-1} -  \frac{\bU_{n-1}\bw_n\bw_n^T}{\|\bw_n\|^2}+  \frac{\by_n \bw_n^T}{\|\by_n\|\|\bw_n\|}$$
where
$$\by_n = \cos (\theta_n) \frac{\bp_n}{\|\bp_n\|}  +  \sin(\theta_n) \frac{\br_n}{\|\br_n\|} \;. $$
This form makes it clear that GROUSE is simply replacing a direction in the current subspace estimate, $\bp_n = \bU_{n-1} \bw_n$, with a new vector $\by_n$ that is a linear combination of $\bp_n$ and the residual vector $\br_n$. This of course makes $\by_n$ orthogonal to the rest of $\bU_{n-1}$, which is why $\bU_n$ will necessarily also have orthogonal columns.

We note that, if the step size is not given, one can use the step size prescribed in \eqref{grousestep}. This step size maximizes the per-iteration improvement of the algorithm in a greedy way, but can therefore be susceptible to noise. For example, with fully observed data, this greedy step size will replace the direction $\bp_n$ in the current iterate $\bU_{n-1}$ with the observed data $\bx_n$. If bounds on the noise variance are known, one can use the noise-dependent step-size given in \cite{zhang2016global}, which decreases the step as the noise floor is reached.

\begin{algorithm}
\caption{GROUSE \cite{balzano2010online}} \label{alg:grouse}
\begin{algorithmic}[1]
\STATE{Given $\bU_0$, an $d \times k$ orthonormal matrix, $0<k<d$;}
\STATE{Optional input: Step size scheme $\eta_n>0$;}
\STATE{Set $n:=1$;}
\REPEAT 
\STATE{Define $\bw_n := \argmin_{\bw} \|\mathcal P_{\Omega_n} \left(\bx_n - \bU_{n-1} \bw\right)\|_2^2$;} 
\STATE{Define $\bp_n := \bU_{n-1} \bw_n$; 
$$\br_n(i) := \left\{ \begin{matrix} \bx_n(i) - \bp_n(i) & \text{if}\;\Omega_n(i) = 1 \\ 0 & \text{otherwise} \end{matrix}\right. \;.$$}
\IF{$\eta_n$ given}
\STATE{Set $\theta_n = \eta_n \|\br_n\| \|\bp_n\|$.}
\ELSE
\STATE{Set \begin{equation}\label{grousestep} \theta_n = \arctan \left( \frac{\|\br_n\|}{\|\bp_n\|}\right). \end{equation}}
\ENDIF
\begin{align}
\bU_n = \bU_{n-1} &+  \left(\cos (\theta_n)-1 \right) \frac{\bp_n}{\|\bp_n\|} \frac{\bw_n^T}{\|\bw_n\|} \nonumber \\ 
&+  \sin(\theta_n) \frac{\br_n}{\|\br_n\|} \frac{\bw_n^T}{\|\bw_n\|} \;. \label{eq:gpupdate} 
\end{align}
\STATE{$n:=n+1$;}
\UNTIL{termination}
\end{algorithmic}
\end{algorithm}

In a follow-up work, Balzano \emph{et al.} describe SAGE GROUSE  \cite{balzano2013grouse, kennedy2016online}, which was derived in the context of \cref{alg:mdisvd}. SAGE GROUSE replaces $\bS_{n-1}$ with an identity matrix the same size as $\bS_{n-1}$, which makes the algorithm completely agnostic to singular values or the relative weight of singular vectors that have been learned. This can be considered as yet another way of modeling uncertainty in the singular values learned thus far in a streaming context. SAGE GROUSE has been proven to be equivalent to the GROUSE gradient algorithm for a given step size \cite{balzano2013grouse}, showing that indeed the distinction of ``algebraic" and ``geometric" algorithms is not fundamental.  


\begin{remark}[SNIPE]
A block variant of GROUSE was presented in \cite{eftekhari2016snipe}, called Subspace Navigation via Interpolation from Partial Entries (SNIPE). This algorithm partitions the input into blocks and for each block optimizes a subspace to fit the observed entries on that block but remain close to the previous subspace estimate. 
\end{remark}

\subsection{PAST} \label{sec:past}

The Projection Approximation Subspace Tracking (PAST) is proposed by Yang \cite{yang1995projection, yang1996asymptotic} for subspace tracking with full data, which is described in \cref{alg:past}. PAST optimizes the following function at time $n$ without constraining $\bU$ to have orthogonal columns:
\begin{align}
\bU_n &= \argmin_{\bU \in\mathbb{R}^{d\times k}} \sum_{\ell=1}^n \lambda^{n-\ell} \|\bx_\ell-\bU \bU^T\bx_\ell\|_2^2\label{exact-past} ,
\end{align}
where prior observations are discounted by a geometric factor $0\ll\lambda\leq 1$. The name ``projection approximation'' comes from the fact that the projection onto the subspace $\bU$ is approximated by $\bU\bU^T$, without the constraint $\bU^T\bU=\bI_k$. This sum is further approximated by replacing the second $\bU$ in \eqref{exact-past} by $\bU_{\ell-1}$, yielding
\begin{equation}\label{approx-past}
\bU_{n}   = \argmin_{\bU\in\mathbb{R}^{d\times k}} \sum_{\ell=1}^n \lambda^{n-\ell} \|\bx_{\ell}-\bU\bU_{\ell-1}^T\bx_\ell \|_2^2.
\end{equation}
Let the coefficient vector be $ {\bw}_\ell=\bU_{\ell-1}^T\bx_\ell$, then \eqref{approx-past} can be rewritten as
\begin{align}
\bU_n &= \argmin_{\bU\in\mathbb{R}^{d\times k}} \sum_{\ell=1}^n \lambda^{n-\ell} \|\bx_\ell-\bU {\bw}_\ell \|_2^2,
\end{align}
whose solution can be written in a closed-form and efficiently found via recursive least-squares. The PAST algorithm has a computational complexity of $\mathcal{O}(dk)$. PAST has been very popular due to its efficiency, and it has been extended and modified in various ways \cite{abed2000fast,gustafsson1998instrumental,badeau2005fast}.

\begin{algorithm}[t]
\caption{PAST \cite{yang1995projection}} \label{alg:past}
\begin{algorithmic}[1]
\STATE{Given $\bU_0\in \mathbb{R}^{d \times k}$, $\bm{R}_0 =\delta\bI_k$; }
\STATE{Set $n:=1$;}
\REPEAT 
\STATE{Define $ {\bw}_n :=   \bU_{n-1}^T \bx_n $;} 
\STATE $\beta_n =1+\lambda^{-1} {\bw}_n^T \bR_{n-1}   {\bw}_n$,
\STATE $\bv_n =\lambda^{-1} \bR_{n-1}   {\bw}_n$,
\STATE $ \bR_n  =\lambda^{-1} \bR_{n-1}  -   (\beta_n)^{-1} \bv_n \bv_n^T$, 
\STATE $\bU_n =\bU_{n-1}+ (\bx_{n}- \bU_{n-1}  {\bw}_n^T )\bR_n  {\bw}_n$.

\STATE{$n:=n+1$;}
\UNTIL{termination}
\end{algorithmic}
\end{algorithm}

\subsection{PETRELS}
The PETRELS algorithm, proposed in \cite{chi2013petrels}, can be viewed as a modification of the PAST algorithm to handle missing data, which is summarized by Algorithm~\ref{PETRELS}.
PETRELS optimizes the following function at time $n$ without constraining $\bU$ to have orthogonal columns:
\begin{equation}\label{petrels_loss}
\bU_n = \argmin_{\bU\in\mathbb{R}^{d\times k}} \sum_{\ell=1}^n \lambda^{n-\ell} \min_{\bw_\ell \in\mathbb{R}^k} \|\mathcal{P}_{\Omega_\ell}(\bx_\ell-\bU\bw_{\ell})\|_2^2. 
\end{equation}

At each time $n$, PETRELS alternates between coefficient estimation and subspace update. We first estimate the coefficient vector by minimizing the projection residual using the previous subspace estimate:
\begin{align} \label{coef}
\bw_n & =  \argmin_{\bw} \|\mathcal P_{\Omega_n} \left(   \bx_n  - \bU_{n-1} \bw\right)\|_2^2,
\end{align}
where $\bU_0\in\mathbb{R}^{d\times k}$ is a random subspace initialization. The subspace $\bU$ is then updated by minimizing
\begin{equation} \label{parD}
\bU_n =\argmin_{\bU}\sum_{\ell=1}^n\lambda^{n-\ell} \|\mathcal P_{\Omega_\ell} \left( \bx_\ell - \bU  \bw_\ell  \right)\|_2^2,
\end{equation}
where ${\bw}_\ell$, $\ell=1, \cdots, n$ are estimates from \eqref{coef}. The objective function in \eqref{parD} decomposes into a parallel set of smaller problems, one for each row of  $\bU_n=[\bu_n^1,\bu_n^2,\cdots,\bu_n^d]^T$, where $\bu_n^i\in\mathbb{R}^k$. Thus the $i$th row can be estimated by solving
\begin{align} \label{parallel_dm}
\bu_n^i &=\argmin_{\bu\in\mathbb{R}^k}\sum_{\ell=1}^n\lambda^{n-\ell} \Omega_{\ell}(i) ( \bx_{\ell}(i) - \bw_\ell^T\bu)^2 \nonumber \\
& = \bu_{n-1}^i + \Omega_{\ell}(i)  \cdot \left[ \bx_{n}(i)-\bw_n^T\bu_{n-1}^i \right]  \nn \\ 
& \quad\quad \cdot \left( \sum_{\ell=1}^n \lambda^{n-\ell}\Omega_{\ell}(i)  \bw_\ell\bw_\ell^T \right)^{-1}\bw_n
\end{align} 
for $i=1,\cdots, d$. Again, the problem can be solved efficiently via recursive least-squares. Moreover, PETRELS can be made very efficient by parallelizing the implementation of \eqref{parallel_dm}.

\begin{algorithm}[t]
\caption{PETRELS \cite{chi2013petrels}}\label{PETRELS}
\begin{algorithmic}[1]
\STATE{Given $\bU_0=[\bu_0^1,\bu_0^2,\cdots,\bu_0^d]^T$, and $\bR_0^i =\delta\bI_k$, $\delta>0$ for all $i=1, \cdots, d$.}
\STATE{Set $n:=1$;}
\REPEAT 
\STATE{Define $ {\bw}_n :=  \argmin_{\bw} \|\mathcal P_{\Omega_n} \left(   \bx_n  - \bU_{n-1} \bw\right)\|_2^2$ ;} 
\FOR{$i =1, \cdots, d$}
\STATE $\beta_n^i =1+\lambda^{-1}\bw_n^T \bR_{n-1}^i  {\bw}_n$,
\STATE $\bv_n^i =\lambda^{-1} \bR_{n-1}^i  {\bw}_n$,
\STATE $\bR_n^i  =\lambda^{-1} \bR_{n-1}^i  - \Omega_n(i)  \bv_n^i(\bv_n^i)^T/ \beta_n^i$, 
\STATE $\bu_n^i =\bu_{n-1}^i+\Omega_n(i) \left[\bx_{n}(i)- {\bw}_n^T\bu_{n-1}^i\right] \bR_n^i {\bw}_n$.
\ENDFOR

\STATE{$n:=n+1$;}
\UNTIL{termination}

\end{algorithmic}

\end{algorithm}
Both PAST and PETRELS can be regarded as applying second-order stochastic gradient descent \cite{bottou2010large} to the loss function, and each step of the update is approximately a Newton step. Therefore, it is expected that the algorithm will converge quadratically when it is close to the optimal solution. Several algorithms can be developed along similar lines of PETRELS, where the loss function is revised to include regularization terms on the Frobenius norms of the subspace $\bU$ and the weight vector $\bw_n$, which we refer the readers to \cite{mardani2015subspace,feng2013online}.


\section{Performance Analysis}
\label{sec:analysis}

In this section we will describe general analysis methodologies  as well as specific theoretical results for characterizing the performance of the aforementioned streaming PCA and subspace tracking algorithms. 

To carry out the analysis, we need to make assumptions on how the data are generated. A popular approach that has been taken in the literature is to assume that each data vector is generated according to the following ``spiked model'' \cite{johnstone2001distribution}:
\begin{equation}\label{spike}
\bx_n = \bU^{\ast} \ba_n + \sigma \bepsilon_n,
\end{equation}
where $\bU^\ast$ is a deterministic $d \times k$ orthogonal matrix, $\ba_n$ is a random signal vector with covariance matrix $\bSigma_a$, and $\bepsilon_n$ is the noise vector. For simplicity, we assume that the covariance matrix of $\bepsilon_n$ is the identity matrix $\bI_d$, and we use $\sigma$ to denote the noise level. This model arises in applications such as array signal processing, where $\bU^\ast$ is the ``steering matrix'' and $k$ denotes the number of targets to be tracked by the array. The generative model \eqref{spike} can also be seen as a special case of the probabilistic model described in Section~\ref{sec:batch}, since $\mathbb{E}[\bx_n\bx_n^T]=\bU^{\ast}\bSigma_a\bU^{\ast T}+ \sigma^2 \bI_d$ is a sum of an exact low-rank matrix and a full-rank identity matrix. In the missing data case studied in this paper, only a subset of the coordinates of $\bx_n$ are observed. Thus, the actual measurement is $\by_n = \mathcal{P}_{\Omega_n}(\bx_n)$ as in \eqref{eq:mdoperator}, where $\mathcal{P}_{\Omega_n}$ denotes the projection operator onto an observation mask, $\Omega_n\in\{0,1\}^d$. The $i$th entry of $\bx_n$ is observed if and only if $\Omega_n(i)=1$. For simplicity, we shall assume that $\set{\Omega_n(i)}$ is a collection of i.i.d. binary random variables such that
\begin{equation}
\label{Omega}
\mathbb{P}(\Omega_n(i) = 1) = \alpha,
\end{equation}
for some constant $\alpha \in (0, 1)$.

\subsection{Classical Asymptotic Analysis}

Historically, the first analysis of subspace tracking algorithms was done in the asymptotic regime (see, \emph{e.g.}, \cite{yang1996asymptotic,chen1998global}), where the algorithms are shown to converge, in the small step size limit, to the solution of some deterministic Ordinary Differential Equations (ODEs). 

To understand the basic ideas underlying such analysis, we note that the essence of almost all the online algorithms described in Section~\ref{sec:algebraic_methods} and Section~\ref{sec:geometric} is a stochastic recursion of the form
\begin{equation}\label{recursion}
\bU_{n} = \bU_{n-1} + \eta_n Q(\bU_{n-1}, \bx_n, \Omega_n).
\end{equation}
Here, $\bU_n$ is the estimate at time $n$; $Q(\cdot, \cdot, \cdot)$ is some nonlinear function of the previous estimate $\bU_{n-1}$, the new complete data vector $\bx_n$, and its observation mask $\Omega_n$; and $\eta_n$ is the step size (\emph{i.e.}, the learning rate). For example, Krasulina's method given in \eqref{Krasulina} is just a special case of \eqref{recursion} [with $\Omega_n(i) \equiv 1$]. When the step size $\eta_n$ is small, we can perform Taylor's expansion (with respect to $\eta_n$) on the recursion formulas of Oja's method \eqref{Oja} and GROUSE \eqref{eq:gpupdate}, and show that these two algorithms can also be written in the form of \eqref{recursion} after omitting higher-order terms in $\eta_n$.

Under the statistical model \eqref{spike} and \eqref{Omega}, the general algorithm \eqref{recursion} is simply a Markov chain with state vectors $\bU_{n} \in \R^{d \times k}$. The challenge in analyzing the convergence of \eqref{recursion} comes from the nonlinearity in the function $Q(\cdot, \cdot, \cdot)$. In the literature, a very powerful analytical tool is the so-called ODE method. It was introduced to the control and signal processing communities by Ljung \cite{Ljung:1977} and Kushner \cite{Kushner:1977} in the 1970s, and similar approaches have an even longer history in the literature of statistical physics and stochastic processes (see, \emph{e.g.}, \cite{Sznitman:1991, Billingsley:1999} for some historical remarks).

The basic idea of the ODE method is to associate the discrete-time stochastic process \eqref{recursion} with a continuous-time \emph{deterministic} ODE. Asymptotically, as the step size $\eta_n \to 0$ and the number of steps $n \to \infty$, the process \eqref{recursion} can be shown to converge to the solution of an ODE. Specifically, we let the step sizes be such that 
\[
\sum_{n =1}^\infty \eta_n = \infty \ \text{and}\ \sum_{n =1}^\infty \eta_n^2 < \infty.
\]
For example, a popular choice is $\eta_n = c / n$ for some $c > 0$. By defining $t_n = \sum_{\ell \le n} \eta_\ell$ as the ``fictitious'' time, we can convert the discrete-time process $\bU_n$ to a continuous-time process $\bU_t$ via linear interpolation:
\begin{equation}\label{embedding}
\bU_t = \bU_{n-1} + \frac{t - t_{n-1}}{t_{n}-t_{n-1}} (\bU_{n} - \bU_{n-1}), \;   t_{n-1} \le t \le t_{n}.
\end{equation}
Under certain regularity conditions on the function $Q(\cdot, \cdot, \cdot)$, one can then show that, as $t \to \infty$, the randomness in the trajectory of $\bU_t$ will diminish and $\bU_t$ will converge to the deterministic solution of an ODE \cite{Ljung:1977,Kushner:1977}. 

Although a rigorous proof of the above convergence is technical, the limiting ODE, if the convergence indeed holds, can be easily derived, at least in a non-rigorous way. To start, we can rewrite \eqref{recursion} as
\begin{equation}\label{pre_limit}
\frac{\bU_{n} - \bU_{n-1}}{\eta_n} = \mathbb{E}_{\bx_n, \Omega_n \vert \bU_{n-1}} [Q(\bU_{n-1}, \bx_n, \Omega_n)] + m_n,
\end{equation}
where $\mathbb{E}_{\bx_n, \Omega_n \vert \bU_{n-1}}[\cdot]$ denotes the conditional expectation of the ``new information'' $\bx_n, \Omega_n$ given the current state $\bU_{n-1}$, and $m_n $ captures the remainder terms. From the construction of $\bU_t$ in \eqref{embedding}, the left-hand side of \eqref{pre_limit} is equal to $(\bU_{t_{n-1} + \eta_n} - \bU_{t_{n-1}}) / \eta_n$, which converges to $\frac{\dif}{\dif t} \bU_t$ since the step size $\eta_n \to 0$. Moreover, one can show that the remainder $m_n$ is of order $o(1)$. It follows that we can write the limit form of \eqref{pre_limit} as an ODE
\begin{equation}\label{ODE_classical}
\frac{\dif}{\dif t} \bU_t = h(\bU_t),
\end{equation}
where $h(\bU_t) = \mathbb{E}_{\bx, \Omega \,\vert\, \bU} [Q(\bU, \bx, \Omega)]$.


The ODE approach is a very powerful analysis tool. By studying the fixed points of the limiting dynamical system in \eqref{ODE_classical}, we can then draw conclusions about the convergence behavior of the original stochastic process \eqref{recursion}. This approach was taken in \cite{yang1996asymptotic}, where the author used an ODE analysis to show that the PAST algorithm \cite{yang1995projection} globally converges to the target signal subspace $\bU^\ast$ with probability one. This result was later adapted in \cite{chi2013petrels} to analyze PETRELS for the fully observed case.

\subsection{Asymptotic Analysis in the High-Dimensional Regime}

Despite its versatility and strong theoretical value, the above classical asymptotic approach has several limitations: First, the analysis requires the step size $\eta_n$ to tend to zero as $n \to \infty$. While using a decreasing sequence of step sizes $\eta_n$  helps the stochastic algorithm to converge to the globally optimal solution, it is not a good strategy for applications where the target low-dimensional subspace can be time-varying. In that scenario, a small but fixed step size is often more preferable, as it would make the algorithms more nimble in tracking the changing subspace. Second, the classical asymptotic analysis leads to an ODE with $\mathcal{O}(d)$ variables. In modern applications, the number of variables, \emph{i.e.}, $d$ can be very large, making it less practical to numerically solve the ODE.

In what follows, we briefly review a different asymptotic analysis approach \cite{WangL:16, WangEL:17, WangML:17, WangEL:18} that addresses the above problems. For simplicity, we present the underlying idea using the example of Oja's method \eqref{Oja} for learning a one-dimensional subspace using full data, although the same approach applies to the general rank-$k$ case with missing data. 

When $k = 1$, the orthogonalization operator $\Pi$ in \eqref{Oja} is just a normalization, and thus the update rule can be simplified as
\begin{equation}\label{Oja_1d}
\bU_{n} = \frac{\bU_{n-1} + \eta_n \bx_n\bx_n^T\bU_{n-1}}{\norm{\bU_{n-1} + \eta_n \bx_n\bx_n^T\bU_{n-1}}}.
\end{equation}

This stochastic process is a Markov chain in $\R^d$, where the dimension $d$ can be large. To reduce the underlying dimension of the system we need to analyze, we note that the quality of the estimate $\bU_n$ can be fully captured by a scalar quantity
\begin{equation}\label{cos_sim}
s_n \bydef \frac{\bU_n^T \bU^\ast}{\norm{\bU_n} \norm{\bU^\ast}}.
\end{equation}
Clearly, $s_n \in [-1, 1]$, with $s_n = \pm 1$ indicating perfect alignment of $\bU^\ast$ and $\bU_n$. In what follows, we refer to $s_n$ as the \emph{cosine similarity}. 

Substituting \eqref{Oja_1d} and \eqref{spike} into \eqref{cos_sim}, we get a recursion formula for the cosine similarity:
{\small
\begin{align}
&s_{n} = \nonumber\\
&\frac{s_{n-1} + \eta_n(a_n + \sigma p_n)(a_n s_{n-1} + \sigma q_n)}{\big(1 + \eta_n(a_n s_{n-1} + \sigma q_n)^2[2 + \eta_n(a_n^2 + \sigma^2 \norm{\bepsilon_n}^2 + 2 \sigma a_n p_n)]\big)^{\frac{1}{2}}},\label{cos_recursion}
\end{align}}%
where $a_n, \bepsilon_n$ are the signal and noise vector in the generating model \eqref{spike}, respectively, and $p_n \bydef  \bepsilon_n^T \bU^\ast$ and $q_n \bydef \bepsilon_n^T \bU_{n-1}$. The expression \eqref{cos_recursion} might appear a bit complicated, but the key observation is the following: If the noise vector $\bepsilon_n$ is drawn from the normal distribution $\mathcal{N}(\bm{0}, \bI_d)$, then it follows from the rotational symmetry of the multivariate normal distribution that
\[
\mathbb{P}(p_n, q_n \vert s_{n-1}) \sim \mathcal{N}\Big(0, \begin{bmatrix}1 & s_{n-1}\\s_{n-1} & 1\end{bmatrix}\Big).
\]
In other words, given $s_{n-1}$, the two random variables $p_n$ and $q_n$ are joint normal random variables whose distribution is a function of $s_{n-1}$. Consequently, the recursion \eqref{cos_recursion} from $s_{n-1}$ to $s_{n}$ forms a one-dimensional Markov chain. Note that this exact Markovian property relies on the assumption that the noise vector $\bepsilon_n$ be normally distributed. However, due to the central limit theorem, we can show that this property still holds asymptotically, when the underlying dimension $d$ is large and when the elements of $\bepsilon_n$ are independently drawn from more general distributions with bounded moments \cite{WangEL:18}. Moreover, these arguments can be generalized to the missing data case, provided that the subsampling process follows the probabilistic model in \eqref{Omega}.

Further analysis shows that, by choosing the step size $\eta_n = \tau / d$ for some fixed $\tau > 0$, we can apply the similar ODE idea used in the classical asymptotic analysis to obtain a deterministic, limit process for the cosine similarity. More specifically, we can show that a properly time-rescaled version of $s^{(d)}(t) \bydef s_{\lfloor td \rfloor}$ will converge weakly, as $d \to \infty$, to a deterministic function that is characterized as the unique solution of an ODE (see \cite{WangL:16, WangEL:17, WangML:17, WangEL:18} for details). 

In \cite{WangEL:17, WangEL:18}, the exact dynamic performance of Oja's method, GROUSE and PETRELS was analyzed in this asymptotic setting. In what follows, we only state the results for $k=1$. For simplicity, we also assume that the covariance matrix of the signal vector $\ba_n$ in \eqref{spike} is $\bSigma_a = \bI_k$. For PETRELS, it turns out that we just need to study two scalar processes: the cosine similarity $s_n$ as defined in \eqref{cos_sim} and an auxiliary parameter
\[
g_n = d R_n \norm{\bU_n}^{-2},
\]
where $R_n$ is the average of the quantities $\set{R_n^i}$ in \Cref{PETRELS}. Accordingly, the parameter $\delta$ in the algorithm also needs to be rescaled such that $\delta = \delta' / d$ for some fixed $\delta' > 0$. By introducing the ``fictitious'' time $t = n / d$, we can embed the discrete-time sequences $s_n, g_n$ into continuous-time as $s^{(d)}(t) = s_{\lfloor t d \rfloor}$ and $g^{(d)}(t) = g_{\lfloor t d \rfloor}$. As the underlying dimension $d \to \infty$, we can show that the stochastic processes $\set{s^{(d)}(t), g^{(d)}(t)}$ converge weakly to the unique solution of the following systems of coupled ODEs:
\begin{equation}\label{PETRELS_ODE}
\begin{aligned}
\frac{\dif s(t)}{\dif t} & =\alpha s(1-s^2)g-\tfrac{\sigma^{2}}{2}(\alpha s^2+\sigma^{2})s g^{2}\\
\frac{\dif g(t)}{\dif t} & =-g^{2}(\sigma^{2}g+1)(\alpha s^2+\sigma^{2})+\mu g,
\end{aligned}
\end{equation}
where $\alpha$ is the probability with which each coordinate of the data vectors can be observed [see \eqref{Omega}], and $\mu > 0$ is a constant such that the discount parameter $\lambda$ in \eqref{petrels_loss} is set to $\lambda=1-{\mu}/{d}$.

Compared to the classical ODE analysis \cite{yang1996asymptotic,chi2013petrels} which keeps the ambient dimension $d$ \emph{fixed} and studies the asymptotic limit as the step size tends to $0$, the ODEs in \eqref{PETRELS_ODE} only involve $2$ variables $s(t)$ and $g(t)$. This low-dimensional characterization makes the new limiting results more practical to use, especially when the dimension is large.

\begin{figure}[t]
\centering
\includegraphics[width=.45\textwidth]{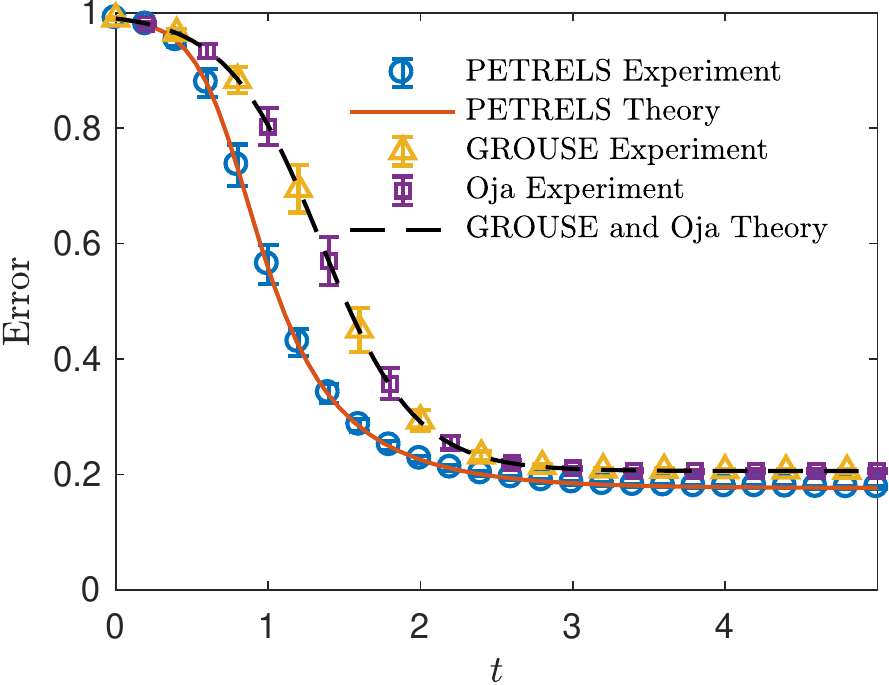}
\hspace{1ex}
\includegraphics[width=0.45\textwidth]{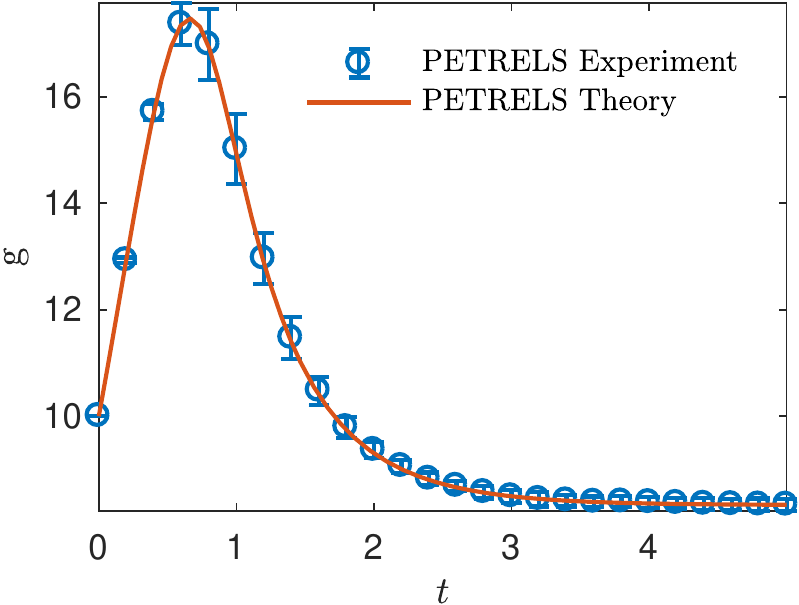}
\caption{\label{fig:f-t}Monte Carlo simulations of the Oja's method, GROUSE and PETRELS v.s. asymptotic predictions obtained by the limiting ODEs given in \eqref{PETRELS_ODE} and \eqref{GROUSE_ODE}. The error is defined as $1-s^2(t)$. The signal dimension is $d = 10^4$, the noise parameter is $\sigma = 0.2$, and the subsampling probability is $\alpha = 0.17$. The error bars shown in the figure correspond to one standard deviation over 50 independent trials. The simulation results also confirm the prediction that Oja's method and GROUSE converge to the same deterministic limit.}
\end{figure}

Similar asymptotic analysis can also be carried out for Oja's method and GROUSE (see \cite{WangEL:18}). Interestingly, the time-varying cosine similarities $s_n$ associated with the two algorithms are asymptotically equivalent, with both converging, as $d \to \infty$, to the solution of a limiting ODE:
\begin{equation}\label{GROUSE_ODE}
\frac{\dif s}{\dif t}=\tau \big(\alpha-\tfrac{\tau \sigma^4}{2}\big)s-\alpha \tau \big(1 + \tfrac{\tau \sigma^2}{2}\big)s^3,
\end{equation}
where $\tau > 0$ is a constant such that the step size parameter $\eta_n$ used in Algorithms~\ref{alg:Oja} and \ref{alg:grouse} is $\eta_n = \tau / d$, and $\alpha$ is again the subsampling probability. Numerical verifications of the asymptotic results are shown in Figure~\ref{fig:f-t}. We can see that the theoretical prediction given by the ODEs \eqref{PETRELS_ODE} and \eqref{GROUSE_ODE} can accurately characterize the actual dynamic performance of the three algorithms.

\begin{figure}[t]
\centering{}\quad\includegraphics[width=0.65\linewidth]{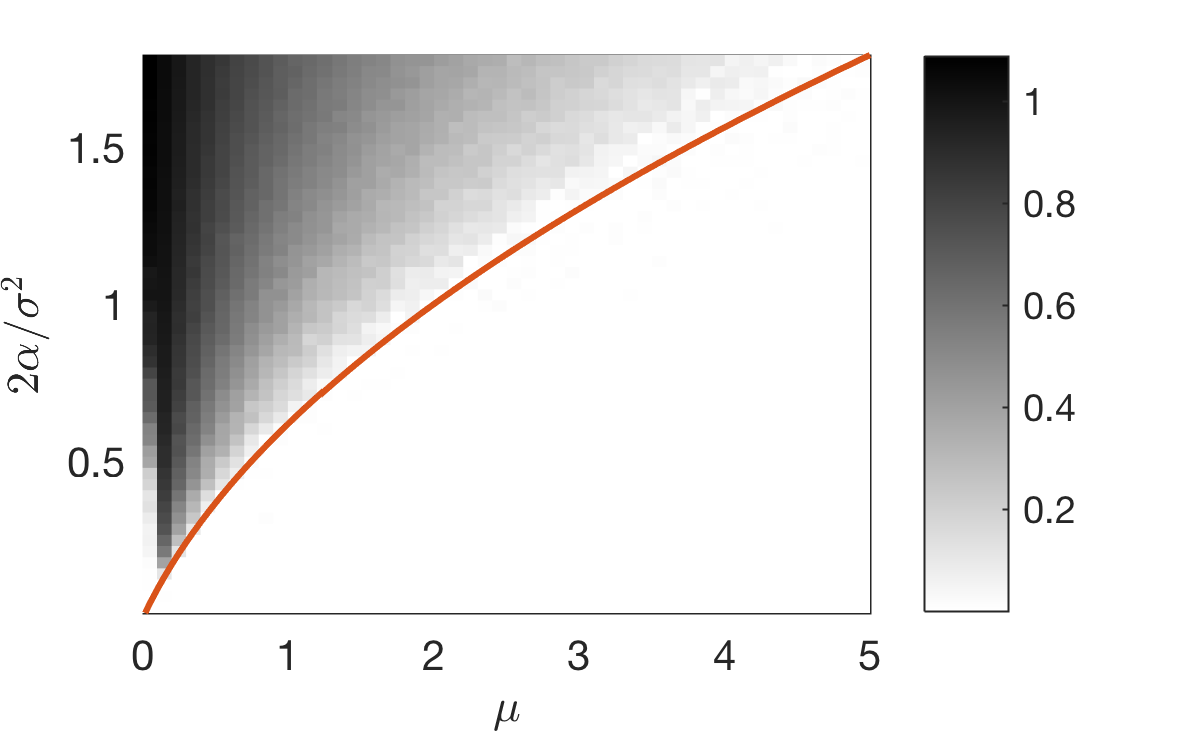}\caption{\label{fig:phase}The grayscale in the figure visualizes the steady-state errors of the PETRELS algorithm corresponding to different values of the noise variance $\sigma^2$, the subsampling ratio $\alpha$, and the discount parameter $\mu$. The red curve is the theoretical prediction given in \eqref{phase_transition} of a phase transition boundary, below which no informative solution can be achieved by the algorithm. The theoretical prediction matches well with numerical results.}
\end{figure}


The convergence behavior of the algorithms can also be established by analyzing the fixed points of the dynamical system associated with the limiting ODEs. For example, by studying the stability of the fixed points of \eqref{PETRELS_ODE} for PETRELS (see \cite{WangEL:18}), one can show that $ \lim_{t \to \infty}s(t)>0$ if only if 
\begin{equation}\label{phase_transition}
\mu< \left(2 \alpha/ \sigma^{2}+1/2 \right)^{2}-1/4,
\end{equation}
where $\alpha \in (0, 1)$ is the subsampling probability and $\sigma > 0$ is the noise level in \eqref{spike}. A ``noninformative'' solution corresponds to $s(t) =0$, in which case the estimate $\bU_n$ and the underlying subspace $\bU^\ast$ are orthogonal (\emph{i.e.}, uncorrelated). The expression in \eqref{phase_transition} predicts a phase transition phenomenon for PETRELS, where a critical choice of $\mu$ (as a function of $\alpha$ and $\sigma$) 
separates informative solutions from non-informative ones. This prediction is confirmed numerically in \Cref{fig:phase}.

\subsection{Finite Sample Analysis}

In addition to the asymptotic analysis described in the previous subsections, there have also been many recent efforts in establishing finite-sample performance guarantees for various streaming PCA algorithms. We begin with analysis in the case of fully observed data vectors. 

One of the earlier works is \cite{mitliagkas2013memory}, where the authors analyze a block variant of Oja's method: within each iteration, multiple sample vectors are drawn, whose empirical covariance matrix is then used in place of $\bx_n \bx_n^T$ in \eqref{Oja_1d}. Under the generative model \eqref{spike}, the authors show that this algorithm can reach accuracy $\norm{\bU_n - \bU^\ast} \le \varepsilon$, in the rank-one case, if the total number of samples is of order
\[
n = \mathcal{O}\left(\frac{(1+3(\sigma+\sigma^2) \sqrt{d})^2 \log(d / \varepsilon)}{\varepsilon^2 \log[(\sigma^2 + 3/4)/(\sigma^2 + 1/2)]}\right),
\]
where $\sigma$ is the noise level in \eqref{spike}. Similar analysis is available for the general rank-$k$ case. Block Oja's methods have also been studied in \cite{hardt2014noisy,balcan2016improved}, but the analysis is done under a model much more general than \eqref{spike}: the data vectors $\bx_n$ are assumed to be drawn i.i.d. from a general distribution on $\R^d$ with zero-mean and covariance $\bSigma$. 

The performance of Oja's original method has also been studied under the above general model. For the rank-1 case, Li \emph{et al.} \cite{li2018near}  established that the error is on the order of $ \mathcal{O}\left(\frac{\lambda_1\lambda_2}{(\lambda_1-\lambda_2)^2}\frac{d\log n}{n}\right)$, which is near optimal up to a logarithmic factor, with the step size $\eta_n =\frac{2\log n}{(\lambda_1-\lambda_2)n}$. Similarly, Jain \emph{et~al.} \cite{jain2016streaming} provide a near-optimal result for Oja's method, with a time-varying learning rate $\eta_n =\frac{1}{(\lambda_1-\lambda_2)(n+n_1)}$, where $n_1$ is some starting time. Other results include Balsubramani \emph{et~al.} \cite{balsubramani2013fast}, Shamir \cite{shamir2016convergence}, Li \emph{et~al.} \cite{li2016rivalry}. Allen-Zhu and Li \cite{allen2016first} provide a near-optimal guarantee for rank-$k$ Oja's method very recently. This paper also contains a comprehensive table that summarizes many recent results.

Most of the existing analysis and performance guarantees in the literature assume that there is a positive gap between $\lambda_k$ and $\lambda_{k+1}$. This eigengap assumption was removed in Shamir \cite{shamir2016convergence} and Allen-Zhu and Li \cite{allen2016first}, where the authors provide sample complexity bounds that are gap-free, \emph{i.e.} they do not require a strictly positive eigengap.

For fully observed data vectors, the global convergence of the GROUSE algorithm was established  in \cite{zhang2016global, zhang2016convergence} under the generative model \eqref{spike} for the special noise-free case, \emph{i.e.}, $\sigma = 0$. Define $\zeta_n := \det\left(\bU^{\ast T} \bU_n \bU_n^T \bU^{\ast} \right) \in [0,1]$ to be the ``determinant similarity\footnote{Note that the rank-$k$ generalization of the aforementioned cosine similarity \eqref{cos_sim} is given as $\|\bU_n^T \bU^\ast\|_F^2$ for $\bU_n$ and $\bU^\ast$ with orthonormal columns. Therefore the two similarity measures are related via the singular values of $\bU_n^T \bU^\ast$.}" between two subspaces, which will be 0 when the subspaces have any orthogonal direction and will be 1 if and only if the two subspaces are equal. Let $\zeta^*<1$ be the desired accuracy of our estimated subspace. 
Initialize the starting point ($\bU_0$) of GROUSE as the orthonormalization of a $d\times k$ matrix with entries being standard normal variables. Then for any $\rho>0$, after 
\[
n \geq \underbrace{\bigg(\frac{2 k^2}{\rho} + 1\bigg)\mu_0 \log (d)}_{n_1} + \underbrace{2 k \log \bigg( \frac{1}{2\rho(1-\zeta^{\ast})}\bigg)}_{n_2} \nn
\]
iterations of GROUSE Algorithm~\ref{alg:grouse},  $ \zeta_n \geq \zeta^\ast$ with probability at least $1-2\rho$,
    where $\mu_0 = 1 + \frac{\log \frac{(1 - \rho')}{C} + k\log (e/k)}{k\log d}$ with $C > 0$ a constant approximately equal to $1$. 
  \label{thm:global} This result is divided into two parts: $n_1$ is the bound on the number of observed vectors (also iterations) required to get to a basin of attraction, and $n_2$ is the number required for linear convergence in the basin of attraction. In practice we see that the bound on $n_1$ is loose but $n_2$ is accurate when discarding dependence on $\rho$ (see \cite{zhang2016global,zhang2016convergence} for more details.)

In contrast to the fully observed case, the literature is much sparser for finite-sample analysis in the missing data case. While a great deal of work has established performance guarantees in the batch setting (\emph{i.e.} the low-rank matrix completion problem \cite{candes2009exact, keshavan2010matrix, recht2011simpler, koltchinskii2011nuclear, davenport2016overview}), the only results in the streaming case with missing data are local convergence results for GROUSE and the work of \cite{desa2015global}, where the authors consider recovering a low-rank positive semidefinite matrix observed with missing data using stochastic gradient descent; while the authors of \cite{desa2015global} did not study streaming PCA in particular, their method could be applied in the streaming data case. 

Bounds on the expected improvement at each iteration for the GROUSE algorithm has been given for the case of noisy data \cite{zhang2016global} and missing or compressed data \cite{balzano2015local, zhang2016convergence}. Under some assumptions of data regularity and subspace incoherence, these convergence results show that the GROUSE algorithm improves $\zeta_{n+1}$ in expectation as follows:
\begin{equation}
  \mathbb{E}\left[\zeta_{n + 1} \big\lvert \bU_n\right] \geq \left(1 + \eta \frac{|\Omega_n|}{d}\frac{1 - \zeta_n}{k}\right)\zeta_n
  \label{eq:expected_rate}
\end{equation}  with high probability, where $\eta \approx 1$ is slightly different for each sampling type \cite{zhang2016global, zhang2016convergence}. The expectation is taken with respect to the generative model \eqref{spike} and the subsampling model \eqref{Omega}. These results for missing or compressed data can be generalized to a local convergence result \cite{balzano2015local, zhang2016convergence}.

In the work of \cite{desa2015global}, the authors propose a step size scheme for general Euclidean stochastic gradient descent with which they prove global convergence results from a randomly generated initialization. Their choice of step size depends on the knowledge of some parameters that are likely to be unknown in practical problems. Without this knowledge, the results only hold with sufficiently small step size that implies very slow convergence.

It remains an important open problem to establish finite sample global performance guarantees in the missing data case for GROUSE and other algorithms such as Oja's and PETRELS.

\section{Numerical Experiments}
\label{sec:comparisons}

We benchmark the performance of several competitive algorithms reviewed in this paper that are able to handle missing data, including GROUSE \cite{balzano2010online}, PETRELS \cite{chi2013petrels}, PIMC \cite{kennedy2014online}, MD-ISVD \cite{kennedy2016online}, Brand's algorithm \cite{brand2002incremental} and Oja's algorithm \cite{oja1982simplified} adapted to the missing data case as proposed in this paper in \cref{alg:Oja}. The code for these simulations is available in Python Jupyter notebooks at \url{http://gitlab.eecs.umich.edu/girasole/ProcIEEEstreamingPCA/}.

\subsection{Simulation Setup}
Let $d=200$ and $k=10$. We select the ground truth subspace as $\bm{U}^* =\text{orth}(\widetilde{\bm{U}})\in\mathbb{R}^{d\times k}$ where $\widetilde{\bm{U}}$ is composed of standard i.i.d. Gaussian entries. The coefficient vectors are generated as $\bm{a}_n \sim\mathcal{N}(\bm{0},\mbox{diag}(\bm{c}))$, where the loading vector $\bm{c}$ is given as  
$$\bm{c}=[1,1,1,1,1,1,1,1,1,1]^T,$$ 
for a well-conditioned system, and
$$\bm{c}= [1,1,1,1,1,0.3,0.3,0.3,0.1,0.1]^T$$
for an ill-conditioned system.
The data vector is generated using \eqref{spike}, where $\bm{\epsilon}_n$ is composed of i.i.d. $\mathcal{N}(0,\sigma^2)$ entries, with the noise level given as $\sigma=[10^{-2},10^{-5},0]$. Each vector is observed with a fixed percent $\alpha=[0.1, 0.5,1]$ of entries selected uniformly at random. We note that the lowest sampling fraction is near the information-theoretic lower bound of the number of uniform random samples that will guarantee matrix reconstruction. The reconstruction error is calculated as the projection error $\|(\bI - P_{\widehat{\boldsymbol{U}}})\bm{U}^*\|_{F}^2$, where $\widehat{\boldsymbol{U}}$ is the estimated (orthogonalized) subspace. All the algorithms are initialized with the same orthogonalized random subspace. We assume all algorithms are given the true rank. Throughout the simulations, we set the discount parameter $\eta=0.98$ in PETRELS and $\beta=0.98$ in Brand's algorithm, and set $\eta=0.5$ in Oja's algorithm. Note that we did a grid search for these parameters using the setting $\sigma=10^{-5}, \alpha=0.5$, and so picking them differently will likely yield different trade-offs in performance for other levels of noise and missing data. Our goal is to illustrate their typical behaviors without claiming relative superiority.

\subsection{Performance Evaluations}
\begin{figure}[H]
\begin{center}
\includegraphics[trim={0 2.8cm 0 2.5cm},clip,height=3.2in,width=0.95\textwidth]{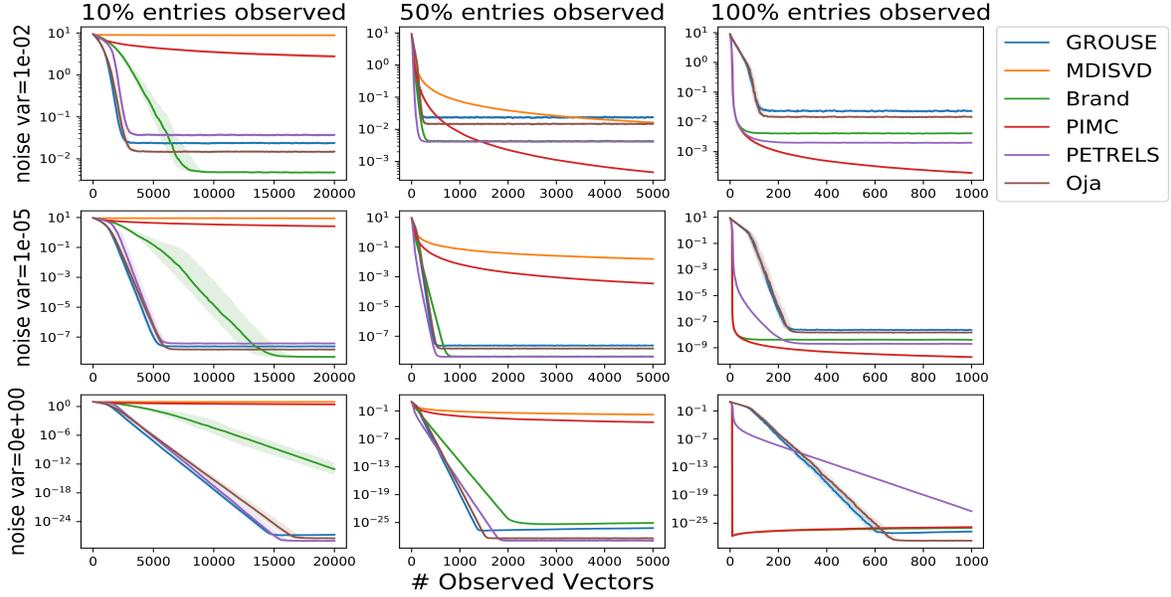} \\
(a) Subspace error versus number of observed vectors in the well-conditioned case \\
\includegraphics[trim={0 2.8cm 0 2.5cm},clip,height=3.2in,width=0.95\textwidth]{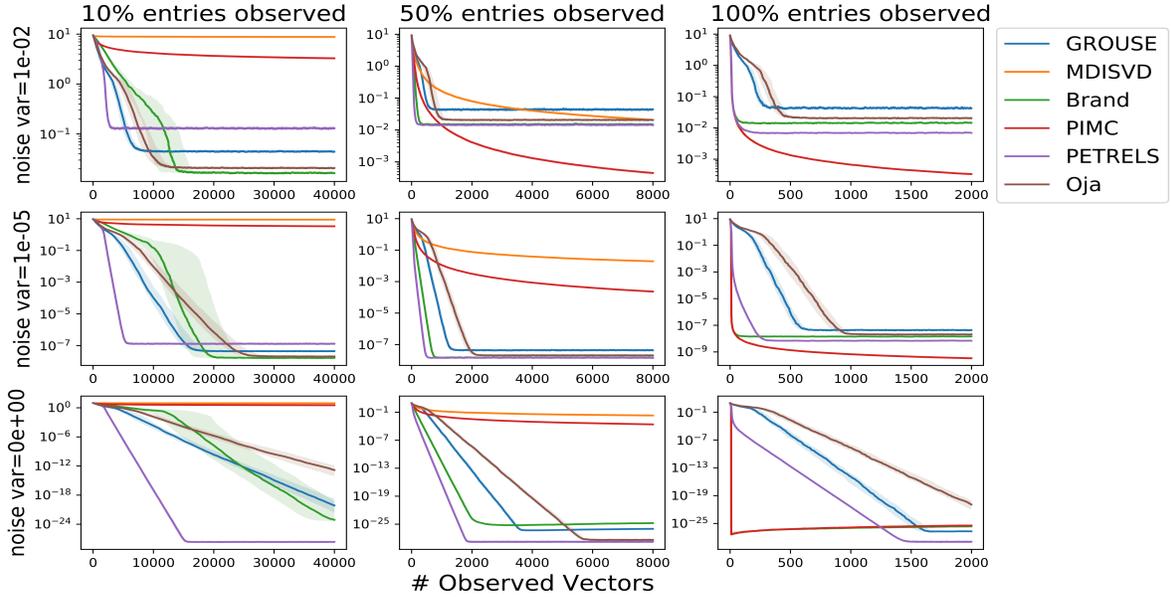} \\
(b) Subspace error versus number of observed vectors in the ill-conditioned case
\end{center}
\caption{Reconstruction error versus the number of observed snapshots for various algorithms in both well-conditioned and ill-conditioned cases for streaming PCA where every data vector is observed once. \vspace{-0.1in}} \label{fig:error_iter}
\end{figure}

We first examine the performance of subspace estimation with streaming observations of a static subspace. Fig.~\ref{fig:error_iter} shows the reconstruction error of the algorithms with respect to the number of snapshots for both the well-conditioned case and the ill-conditioned case. 
For each algorithm, the dark line is the median of 50 runs, and the transparent ribbon shows the 25\% and 75\% quantiles. Note how in most algorithms, the ribbon is nearly invisible, meaning that the algorithm execution with different initialization, generative subspace, random data draw from that subspace, and random missing data patterns have highly consistent behavior, especially as the number of observed entries gets larger. All of the examined algorithms are capable of estimating the subspace after seeing sufficiently many data vectors.
It can be seen that as the fraction of missing data increases, more snapshots are needed for the algorithms to converge.   
Also as the noise level increases, the algorithms converge to a higher error level.

\begin{figure}[H]
\begin{center}
\includegraphics[trim={0 2.8cm 0 2.5cm},clip,height=3.2in,width=0.95\textwidth]{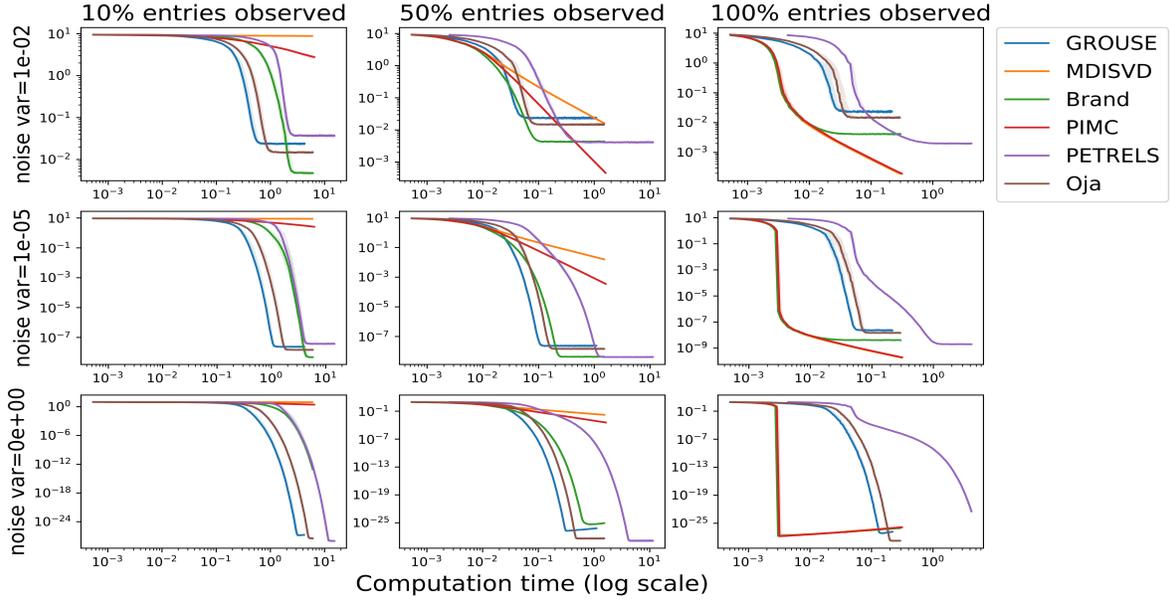} \\
(a) Subspace error versus computation time for the well-conditioned case \\
\includegraphics[trim={0 2.8cm 0 2.5cm},clip,height=3.2in,width=0.95\textwidth]{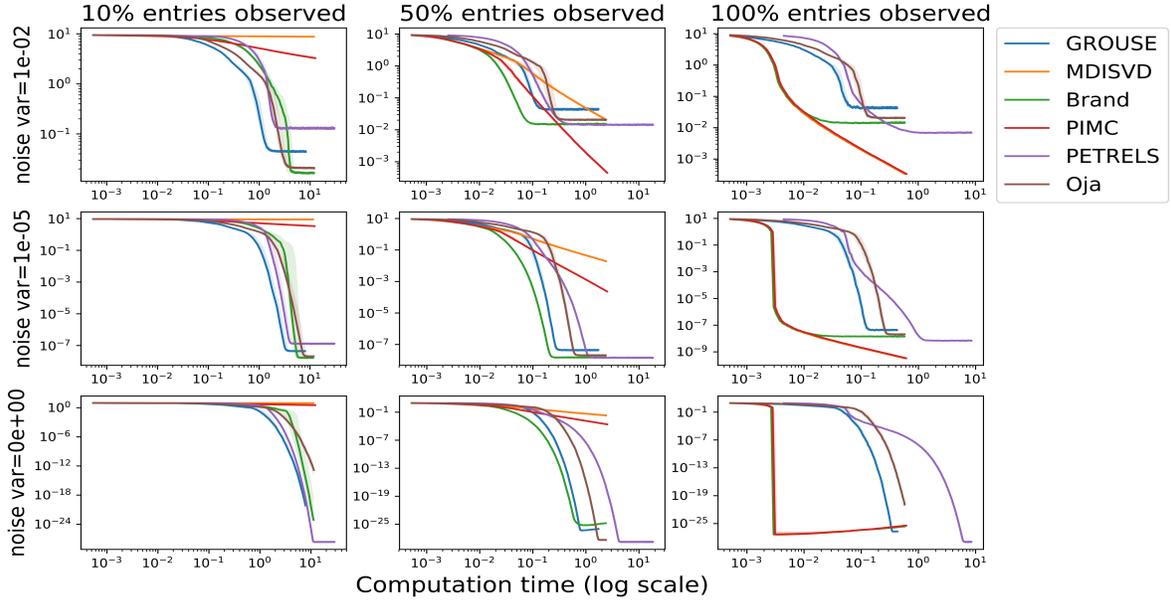}   \\
(b) Subspace error versus computation time for the ill-conditioned case
\end{center}
\caption{Reconstruction error versus the wall-clock time for various algorithms in both well-conditioned and ill-conditioned cases for one-pass streaming PCA. \vspace{-0.1in}} \label{fig:error_time}
\end{figure}

We next show the same plots but as a function of computation time, i.e. assuming that the streaming vectors are available as soon as each algorithm finishes processing the previous vector. Fig.~\ref{fig:error_time} shows the reconstruction error of the algorithms with respect to the wall-clock time for both the well-conditioned case and the ill-conditioned case. For each algorithm, we take the median computation time at each iteration over 50 runs to get values on the x-axis. Then the the dark line is the median error of 50 runs interpolated to those computation times. Again the transparent ribbon shows the 25\% and 75\% error quantiles. We show the time on a log axis since different algorithms take different orders of magnitude computation time. Note that PETRELS can be accelerated by computing in parallel the updates for each row of the subspace, which is not implemented here. Again, while many algorithms converge very quickly with complete data, they may take significantly longer in the presence of missing data.   

\begin{figure}[ht]
\begin{center}
\includegraphics[width=.65\textwidth]{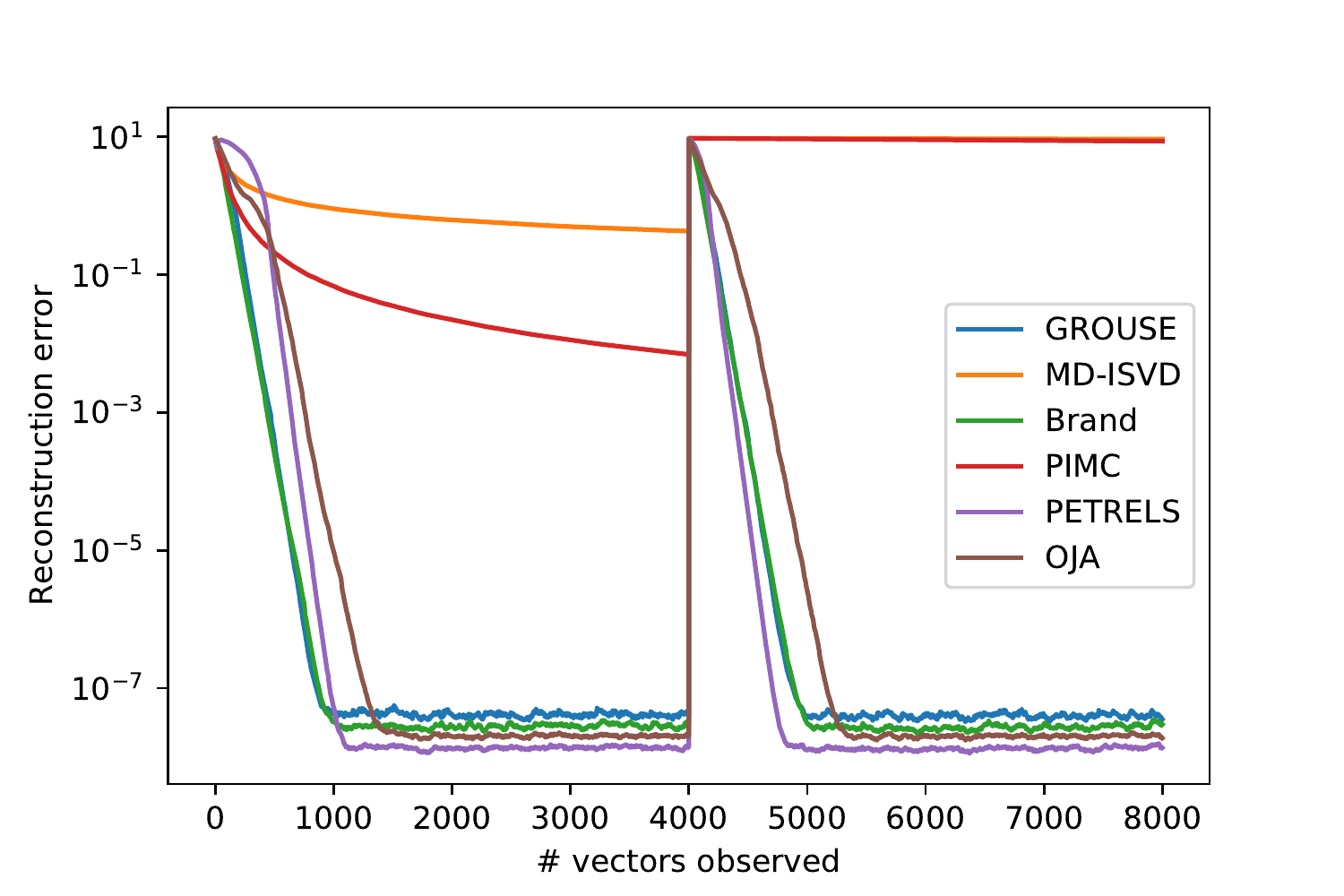} 
\end{center}
\caption{Reconstruction error versus the number of observed snapshots with an abrupt subspace change.} \label{fig:abrupt_change}
\end{figure}

Next, we examine the performance of the algorithms on subspace tracking, where we assume there is an abrupt change in the underlying subspace and the goal is to examine how fast the algorithms are able to track the change. Moreover, we generate the loading vector $\bm{c}$ before and after the same with entries drawn from a uniform distribution in $[0,1]$. The noise level is set as $\sigma=10^{-5}$ and the fraction of observation is set as $\alpha=0.3$. Fig.~\ref{fig:abrupt_change} shows the performance of the algorithms when the subspace changes abruptly at the 4000th snapshot. Note that MD-ISVD and PIMC have poor performance after the subspace change; these algorithms seem to rely more heavily on historical data through their singular value estimates.

\begin{figure}[ht]
\begin{center}
\includegraphics[width=.65\textwidth]{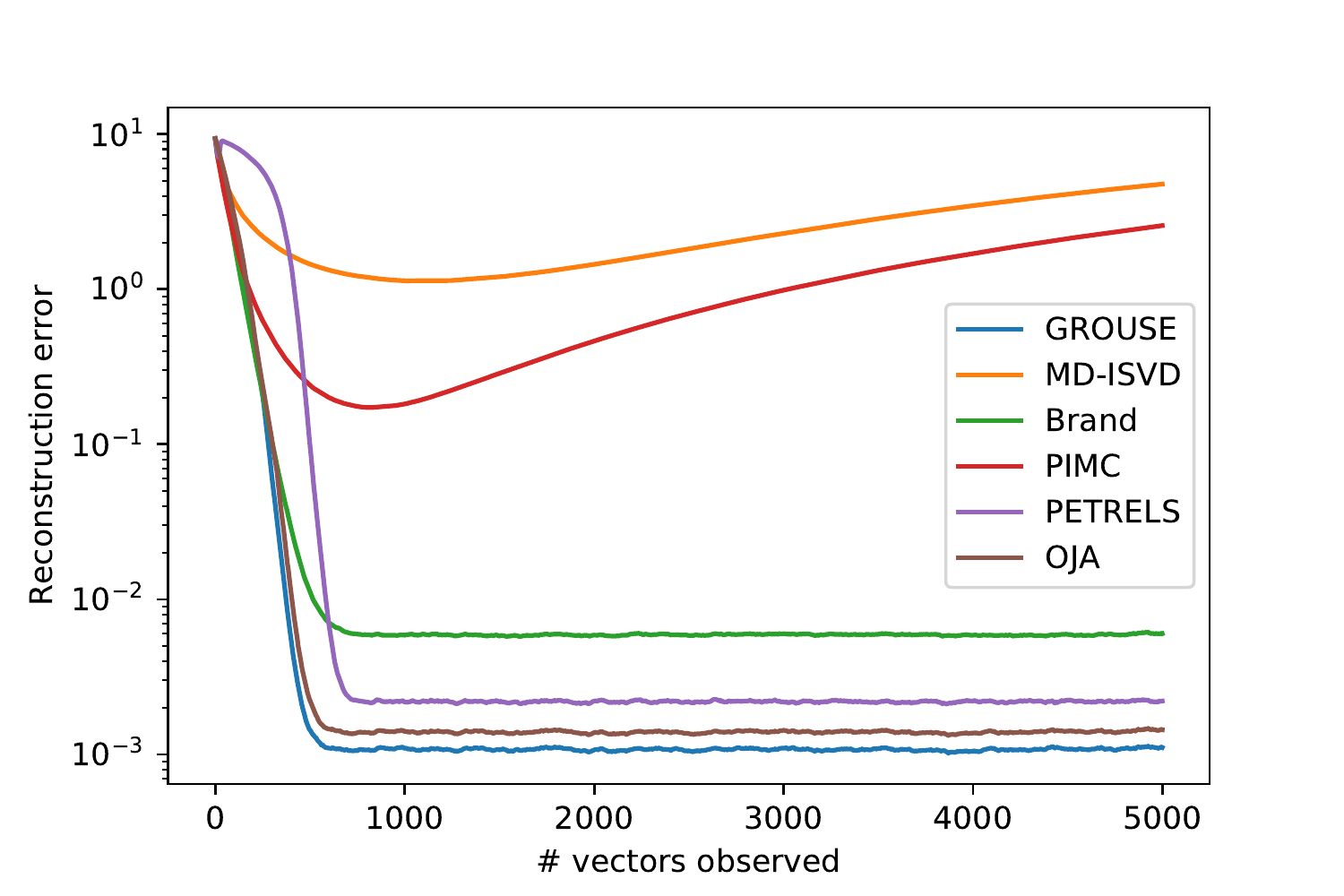} 
\end{center}
\caption{Reconstruction error versus the number of observed snapshots with a subspace that slowly rotates over time.} \label{fig:rotate}
\end{figure}

Finally in a second experiment for subspace tracking, we assume the underlying subspace is slowly rotating over time. We achieve this by initializing a random subspace with orthonormal columns $\bU_0\in\mathbb{R}^{d\times k}$ and sample a skew-symmetric matrix $\bB\in\mathbb{R}^{d\times d}$ with independent, normally distributed entries. We set $\bU_t = \exp(\delta_0 \bB)\bU_{t-1}$ for some small rotation parameter $\delta_0$ and where $\exp$ is the matrix exponential. In our experiments we set $\delta_0 =10^{-5}$. Again, the noise level is set as $\sigma=10^{-5}$ and the fraction of observation is set as $\alpha=0.3$, but here we set $\bc$ to be the all-ones loading vector. The results can be seen in Fig.~\ref{fig:rotate}; notice both that some of the algorithms fail to converge (again MD-ISVD and PIMC), and the rest converge but do not achieve as good accuracy as in the static case. This is to be expected since none of the algorithms is doing a prediction, but instead only tracking based on available data; therefore the constant difference between the true subspace at two consecutive time steps contributes to the noise floor for the algorithms.

\section{Concluding Remarks}\label{sec:final}
In this paper, we have reviewed a variety of streaming PCA and subspace tracking algorithms, focusing on the ones that have a near-linear memory complexity and computational time, appealing convergence rates, and can tolerate missing data. Convergence guarantees based on classical and new asymptotic analysis as well as recent finite-sample non-asymptotic analysis are discussed. While we divide our discussions intro algebraic and geometric approaches, it is important to point out that these approaches are not actually distinct. For example, Oja's is equivalent to Krasulina's, a stochastic gradient descent method, by ignoring second-order terms; GROUSE is equivalent to a version of Incremental SVD that is agnostic to past singular values \cite{balzano2013grouse}; and the PAST algorithm can also be interpreted from both perspectives \cite{hua1999new}. It is an exciting open question to understand the connections between algebraic and geometric methods more generally.

Due to space limitations, we have focused on the problem of estimating a single low-dimensional subspace under the squared loss for streaming data with possibly missing entries, which is most suitable when the noise is modeled as Gaussian. There are many important extensions, motivated by real-world applications, including subspace tracking for non-Gaussian data \cite{shen2017online,wang2018poisson}, tracking a union-of-subspace model \cite{balzano2012k,jiang2016online}, tracking a low-dimensional subspace with multi-scale representations \cite{xie2013change}, subspace tracking in the presence of outliers and corruptions \cite{vaswani2017robust,feng2013online,goes2014robust, he2012incremental,narayanamurthy2017provable,zhan2016online}, adaptive sampling \cite{krishnamurthy2013low}, for an incomplete list. Interested readers are invited to go to the above cited references for details.


\section*{Acknowledgment}
L. Balzano would like to thank her lab and especially Dejiao Zhang and David Hong for their comments and help with the jupyter notebook experiments. Y. Chi would like to thank Myung Cho for his help with the experiments. We also thank the reviewers for their very thoughtful feedback. 
The work of L. Balzano is supported in part by DARPA grant 16-43-D3M-FP-03 and NSF Grant ECCS-1508943.
The work of Y. Chi is supported in part by the grants AFOSR FA9550-15-1-0205, ONR N00014-18-1-2142, NSF CCF-1806154, ECCS-1818571 and NIH R01EB025018. The work of Y. M. Lu is supported in part by the ARO under contract W911NF-16-1-0265 and by NSF under grants CCF-1319140 and CCF-1718698.

\bibliographystyle{IEEEtran}
\bibliography{streamingPCA.bib}

\end{document}